\documentclass[runningheads]{llncs}
\usepackage[T1]{fontenc}
\usepackage{graphicx}
\usepackage{subfigure}
\usepackage[ruled,vlined,linesnumbered]{algorithm2e}
\usepackage{multirow}
\usepackage{amsmath} 
\usepackage{setspace}
\usepackage{booktabs}
\usepackage[misc]{ifsym}

\usepackage{hyperref}
\hypersetup{hidelinks,
	colorlinks=true,
	allcolors=blue,
	pdfstartview=Fit,
	breaklinks=true}

\usepackage{bm}
\usepackage{amsfonts} 
\usepackage{stmaryrd} 

\usepackage{todonotes}
\usepackage{comment}
 
\usepackage{color}
\definecolor{Orange}{rgb}{1,0.5,0}
\definecolor{Red}{rgb}{1,0,0}
\definecolor{Blue}{rgb}{0,0,1}

\begin{document}
\tocauthor{
Ao Zhou, Bin Liu, Jin Wang, Kaiwei Sun, Kelin Liu
}

\toctitle{
AEMLO: AutoEncoder-Guided Multi-Label Oversampling
}
\title{AEMLO: AutoEncoder-Guided Multi-Label Oversampling}

\author{Ao Zhou\inst{1} \and
Bin Liu\inst{1} \Letter \and
Jin Wang\inst{1} \and
Kaiwei Sun\inst{1} \and
Kelin Liu\inst{1}}
\authorrunning{Ao Zhou et al.}
\institute{Key Laboratory of Data Engineering and Visual Computing,\\
Chongqing University of Posts and Telecommunications, China\\
\email{zacqupt@gmail.com, \{liubin, wangjin,sunkw\}@cqupt.edu.cn, Cling798as@gmail.com}}
\maketitle

\begin{abstract}
Class imbalance significantly impacts the performance of multi-label classifiers. Oversampling is one of the most popular approaches, as it augments instances associated with less frequent labels to balance the class distribution. Existing oversampling methods generate feature vectors of synthetic samples through replication or linear interpolation and assign labels through neighborhood information. Linear interpolation typically generates new samples between existing data points, which may result in insufficient diversity of synthesized samples and further lead to the overfitting issue. Deep learning-based methods, such as AutoEncoders, have been proposed to generate more diverse and complex synthetic samples, achieving excellent performance on imbalanced binary or multi-class datasets. In this study, we introduce AEMLO, an AutoEncoder-guided Oversampling technique specifically designed for tackling imbalanced multi-label data. AEMLO is built upon two fundamental components. The first is an encoder-decoder architecture that enables the model to encode input data into a low-dimensional feature space, learn its latent representations, and then reconstruct it back to its original dimension, thus applying to the generation of new data. The second is an objective function tailored to optimize the sampling task for multi-label scenarios. We show that AEMLO outperforms the existing state-of-the-art methods with extensive empirical studies.
\end{abstract}

\section{Introduction}
In the field of multi-label classification (MLC), each instance can belong to multiple labels simultaneously. 
MLC is widely used in various fields, including image annotation\cite{image2}, sound processing \cite{sound}, biology \cite{bio} and text classification  \cite{text1}. 
The issue of class imbalance in multi-label classification has gained prominence recently \cite{mulitilabelimbalancereview}. It is prevalent in real-world MLC problems and significantly affects classifier performance, as many algorithms assume data is balanced. Imbalanced datasets tend to bias learners towards majority labels \cite{MLBOTE}.
\subsection{Research Goal}
Our goal is to address the class imbalance in multi-label datasets through the integration of a deep generative model within an encoder-decoder architecture. 
This strategy seeks to outperform conventional methods, such as sampling with linear interpolation or random replication, by dynamically creating instances that contain richer feature information.

\subsection{Motivation}
In recent years, innovative approaches have been developed to tackle the issue of imbalance in multi-label learning \cite{mulitilabelimbalancereview}, including sampling methods \cite{MLSMOTE,MLSOL}, classifier adaption \cite{SOSHF}, and ensemble techniques \cite{ECCRU3,Ensemble1}. Sampling methods, in particular, aim to balance the dataset before the training phase, offering flexibility and compatibility with any multi-label classifier. 
To ensure effective sampling, several studies have concentrated on identifying specific samples and refining decision boundaries. For example, MLSOL \cite{MLSOL} assigns a higher selecting probability to the sample suffering severe local imbalance. MLBOTE \cite{MLBOTE} refines the boundary samples related to high imbalance labels and employs different sampling strategies. Traditional oversampling techniques often rely on basic linear interpolation or replication for creating feature vectors of synthesized samples, with label vectors typically generated through majority voting or replication.

The Autoencoder (AE) and Generative Adversarial Network (GAN), as exemplary generative models, have shown substantial potential in data generation, restoration, and augmentation \cite{VAE,GAN,BAGAN,VAEsampling1,Deepsmote}. 
An Autoencoder compresses data into a latent space using an encoder and reconstructs it by a decoder. Its objective is to minimize reconstruction errors, enabling efficient feature extraction and noise reduction \cite{AE,VAE}. 
Although Autoencoder and GAN are used to address the imbalance problem and generate minority samples, they primarily cater to single-label datasets and 
face several challenges when applied to multi-label datasets. 
First, AE and GAN require training samples with identical labels (same class in the single-label dataset or same label set in the multi-label dataset). However, in multi-label data, the number of samples with a complete label set is often too limited to effectively train deep learning models.
Secondly, although multi-label datasets can be divided into several binary datasets via the \textit{One vs All} strategy, For each binary dataset, we can learn and reconstruct new feature vectors for multi-label data through end-to-end models, but we can not determine appropriate complete label set for each feature vector.

\begin{figure}[h]
\centering
\includegraphics[width=0.8\textwidth]{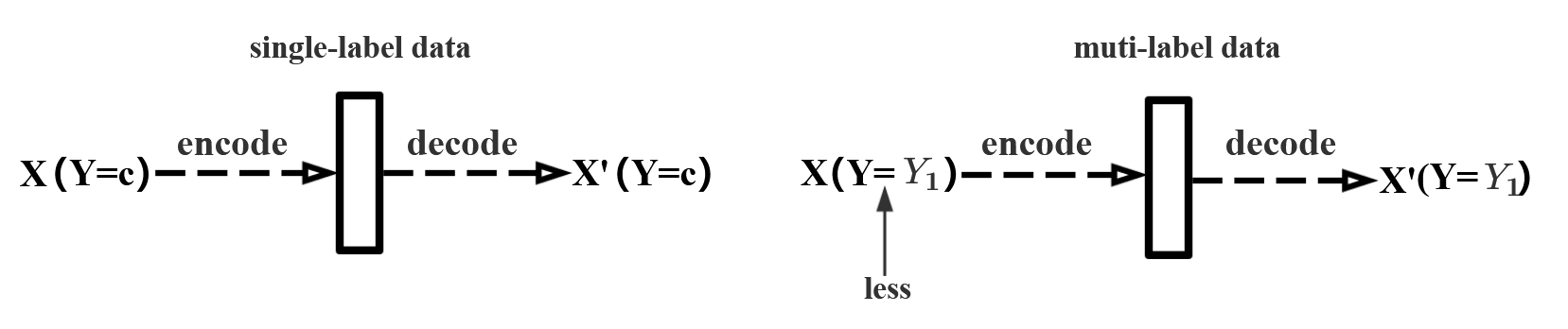}
\caption{Using Autoencoders to train data.}
\label{intro}
\end{figure}

\subsection{Summary}
In this work, we introduce an innovative approach crafted to tackle the class imbalance issue in multi-label datasets named \textbf{A}uto\textbf{E}ncoder-guided \textbf{M}ulti-\textbf{L}abel \textbf{O}versampling (AEMLO). The core of AEMLO's design lies in two essential elements:
\begin{enumerate}
    \item The basic encoder-decoder architecture is designed to encode data into a lower-dimensional space and subsequently reconstruct it, making it suitable for oversampling applications.
    \item A specialized objective function for multi-label imbalance data sampling. 
\end{enumerate}
Our approach incorporates the sampling process into a deep encoder-decoder framework that has been pre-trained, providing a holistic solution for the creation of low-dimensional data representations and synthetic instances through an end-to-end methodology. By augmenting the original training set with instances generated via AEMLO, we further train various traditional multi-label classifiers and conduct comparisons against several multi-label sampling techniques. Experimental results consistently demonstrate the superiority of our method. Our code can be found in \href{https://github.com/CquptZA/AEMLO}{https://github.com/CquptZA/AEMLO}.

\section{Related Work}
\subsection{Multi-Label Classification}
Formally, let \(\mathcal{X} \in \mathbb{R}^d\) represent the \(d\)-dimensional feature space, and let \(L = \{l_1, l_2, \ldots, l_q\}\) denote a set of \(q\) predefined labels. In multi-label classification, our objective is to construct a mapping function \(h: \mathcal{X} \rightarrow L \) based on a given multi-label training dataset \(D = \{(\mathbf{x}_i, \mathbf{y}_i)\}_{i=1}^n\), where each sample \(\mathbf{x}_i \in \mathcal{X}\) is associated with a binary label vector \(\mathbf{y}_i \in \{0, 1\}^q\). Here, \(\mathbf{y}_i\) is a binary vector where each element denotes whether the associated label from \(L\) is relevant (1) or not relevant (0) to \(\mathbf{x}_i\). 

In Multi-Label Classification (MLC), methods are split into three types based on how they handle label correlations. First-order strategies like MLkNN \cite{MLKNN} and BR \cite{BR} treat labels independently, offering simplicity and efficiency. Second-order methods, such as CLR \cite{CLR}, analyze pairwise label correlations for improved interaction understanding. For complex scenarios with intricate label relationships, high-order methods, like RAkEL \cite{RAkEL} and ECC \cite{ECC}, are more effective. RAkEL tackles this by dividing labels into subsets for diverse interaction modeling. ECC sequentially links classifiers, allowing each to learn from the predictions of its predecessors.
\subsection{Multi-Label Imbalance Learning}
Let $N_{j}^{1}$($N_{j}^{0}$) denote the number of instances with "1" ("0") class of label $l_j$. $IRlbl$ and $ImR$ are the two measures to evaluate the imbalance level of individual labels \cite{MLSMOTE,COCOA}. Let $N_{max}^{1}=max\{N_j^1\}_{j=1}^q$ be the number of "1"s in the most frequent label, $IRlbl_j$ and $ImR$ are defined as:
\begin{equation}
IRlbl_{j}=N^{1}_{max}/N^{1}_{j} \quad ImR_{j}=max(N^{1}_{j},N^{0}_{j})/min(N^{1}_{j},N^{0}_{j})
\label{irlblandImR}
\end{equation}
The larger the $IRlbl_j$ and $ImR_j$, the higher the imbalance level of $l_j$. Then, $MeanIR$ calculates the average imbalance ratio ($IRlbl$) across all labels in a dataset, defined by: $\frac{1}{q}\sum_{j=1}^{q}IRlbl_{j}$, where $q$ is the total number of labels. The higher $MeanIR$, the imbalance of the dataset. By considering the $IRlbl$ and $MeanIR$, we can calculate imbalance indicators such as the coefficient of variation of $IRlbl$ ($CVIR$) and concurrency level ($SCUMBLE$) \cite{mulitilabelimbalancereview}.

The imbalanced approaches proposed for MLC can be divided into three categories: sampling methods, classifier adaptation \cite{SOSHF,COCOA,Adaption1}, and ensemble approaches \cite{Ensemble1,ECCRU3}.  
Compared to the other two methods, the sampling method is more universal, as it creates (deletes) instances related to minority (majority) labels to construct a balanced training set that can be used to train any classifier without suffering from bias.
Sampling methods involve undersampling and oversampling techniques. Undersampling reduces the presence of majority labels by either randomly removing instances or employing heuristic approaches to selectively eliminate samples. For example, LPRUS and MLRUS \cite{ML} aim to alleviate imbalances by respectively targeting the most frequent label sets or individual labels for removal. Conversely, oversampling techniques such as LPROS and the MLSMOTE \cite{MLSMOTE} focus on augmenting the dataset with new instances associated with minority labels, either through duplication or the generation of synthetic samples. 
Recent developments include the REMEDIAL \cite{REMEDIAL} method, which adjusts label and feature spaces to lessen label co-occurrence and improve sampling. 
Integrating this method with techniques such as MLSMOTE can further optimize dataset balancing \cite{REMEDIAL-HwR}. 
MLSOL \cite{MLSOL} specifically generates instances to focus on local imbalances in datasets. On the other hand, MLTL \cite{MLTL} refines datasets by removing instances that obscure class boundaries, 
Another notable method, MLBOTE \cite{MLBOTE}, categorizes instances based on their boundary characteristics and applies different sampling rates. 

\subsection{Deep Sampling Method}
Traditional sampling techniques struggle to effectively expand the training set for complex models. This has sparked interest in generative models and their potential to mimic oversampling strategies \cite{deepsampling,Manifoldoversampling}. Utilizing an encoder-decoder setup, artificial instances can be effectively introduced into an embedding space. AE \cite{AE,VAE} and GAN \cite{GAN} have been effectively employed to capture the underlying distribution of data and further applied to generate data for oversampling purposes. 
AE is designed to learn efficient data codings in an unsupervised manner. Essentially, they aim to capture the most salient features of the data by compressing the input into a lower-dimensional latent space and then reconstructing it back to the original dimensionality. The core objective of an AE is defined by the reconstruction error, which quantifies the difference between the original data and its reconstruction. Unlike Variational Autoencoder (VAE), which incorporates the Kullback-Leibler (KL) divergence to regulate the latent space, standard AE relies solely on the reconstruction loss. This encourages the model to develop a compressed representation that retains as much of the original information as possible, enabling the AE to generate reconstructions that are as close to the input. For example. DeepSMOTE \cite{Deepsmote} integrates traditional SMOTE methods into encoding and decoding architectures similar to AE.  
VAE strive to maximize a variational lower bound on the data's log-likelihood. Typically, they are formulated by merging a reconstruction loss with the KL divergence. The KL divergence serves as an indirect penalty for the reconstruction loss, steering the model towards a more faithful replication of the data distribution \cite{VAEsampling1}. By penalizing the reconstruction loss, the model is motivated to refine its replication of the data, thereby enabling it to produce outputs rooted in the input's latent distribution.
GAN has significantly advanced the field of computer vision by framing image generation as a competitive game between a generator and a discriminator network \cite{GANsampling,GANsampling1,BAGAN}. Despite their remarkable achievements, GAN requires the deployment of two separate networks, can encounter training difficulties, and are susceptible to mode collapse \cite{Deepsmote}. 


\section{Multi-Label AutoEncoder Oversampling}
\subsection{Method Description and Overview}
The multi-label AutoEncoder oversampling framework, as described in Algorithm \ref{al:AEMLO}, is divided into the training process and the instance generation phase.
\begin{algorithm}[t]
\KwIn{Original dataset $D=(\mathbf{{X}},\mathbf{{Y}})$, sample count $num$, parameter $\alpha$, $\beta$, latent space dimension $l$}
\KwOut{Balanced training set $D'$}
\tcc{ train the Encoder and Decoder} 
\For{$e \leftarrow 1$ \KwTo $epochs$}{
    \For{$batch(\mathbf{\hat{X}}, \mathbf{\hat{Y}})\gets B$}{
     $\mathbf{F}_{ex}((\mathbf{\hat{X}})),\mathbf{F}_{ey}((\mathbf{\hat{Y}}))$\tcc*[r]{encode batch data to $\mathcal{L}$}
     $\mathbf{F}_{dx}((\mathbf{\hat{X}})),\mathbf{F}_{dy}((\mathbf{\hat{Y}}))$\tcc*[r]{decode batch data from $\mathcal{L}$}
     Define the loss function by Eq \ref{objective function}\;
     Compute gradients and update parameters with Adam\;
    }
    Update $T$ \tcc*[r]{validate and optimize the bipartition threshold for each label}
}
\tcc{ generate instances} 
\While{$num>0$}{
    $\mathbf{x}_s$ $\gets$ select form $M$\tcc*[r]{choose seed instance}
    $\mathbf{F}_{ex}(\mathbf{x}_s)$ \tcc*[r]{encode}
    $\mathbf{x_g} \gets \mathbf{F}_{dx}(\mathbf{F}_{ex}(\mathbf{x_s})) \quad \mathbf{y_g} \gets \mathbf{F}_{dy}(\mathbf{F}_{ex}(\mathbf{x_s}))$\tcc*[r]{decode}
    $\mathbf{y_g}$ $\gets$ $T$\tcc*[r]{rounding}
    D'=D' $\cup$ $(\mathbf{x_g},\mathbf{y_g})$ ;\\
    $num \leftarrow num- 1$ ;\\
}
\KwRet{$D'$} 
\caption{Using Encoder and Decoder for Multi-Label Sampling}
\label{al:AEMLO}
\end{algorithm}

\begin{figure}[!h]
\centering
\includegraphics[width=0.5\textwidth]{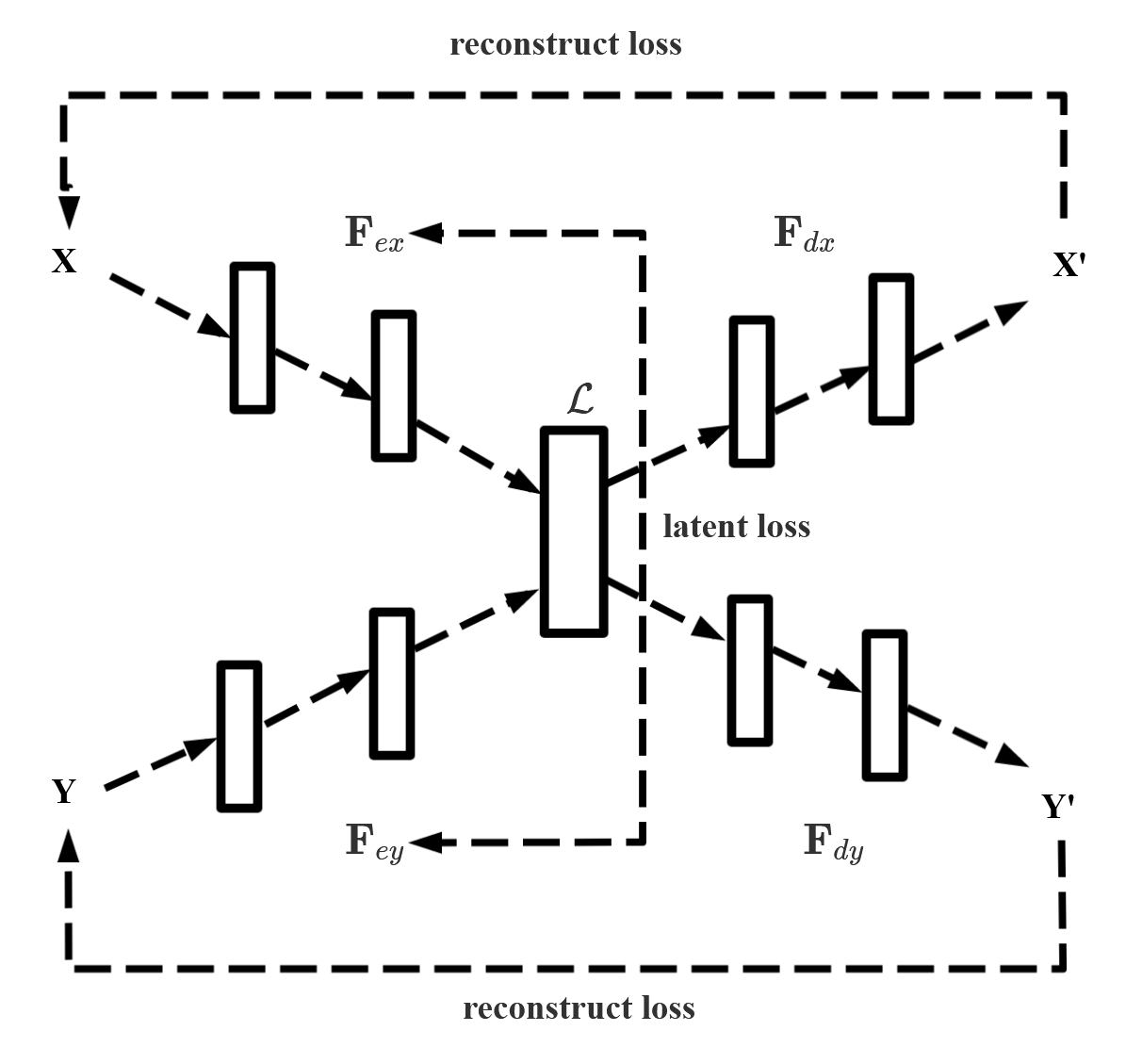}
\caption{The architecture of the proposed autoencoder learns the latent space $\mathcal{L}$ through the function of $\mathbf{F}_{ex}$ and $\mathbf{F}_{ey}$, and decouples $\mathcal{L}$ through $\mathbf{F}_{dx}$ and $\mathbf{F}_{dy}$.}
\label{fig:model}
\end{figure}

In the training process, as shown in Figure \ref{fig:model}, the model is designed to learn and optimize four distinct mapping functions:
the feature encoding function $\mathbf{F}_{ex}$, label encoding function $\mathbf{F}_{ey}$, feature decoding function $\mathbf{F}_{dx}$, and label decoding function $\mathbf{F}_{dy}$. 
The model is trained end-to-end with mini-batches and the Adam optimizer, where batch size $n$ encompasses the feature vector $\mathbf{x}_i$ and binary label vector $\mathbf{y}_i$ of the $i$-th sample, respectively. The matrices $\mathbf{X}$ and $\mathbf{Y}$ aggregate the input features and labels for all samples in the batch. 
The framework ingests a feature matrix $\mathbf{X}$ and its corresponding label matrix $\mathbf{Y}$, aiming to output reconstructed versions of $\mathbf{X'}$ and $\mathbf{Y'}$. Meanwhile, The other goal of our model is to identify an optimal latent space $\mathcal{L}$, where the Deep Canonical Correlation Analysis (DCCA) component \cite{DCCA} enhances the correlation between $\mathbf{X}$ and $\mathbf{Y}$. 
Therefore, the model's objective function is defined as:
\begin{equation}
\Theta = \min_{\mathbf{F}_{ex}, \mathbf{F}_{ey}, \mathbf{F}_{dx},\mathbf{F}_{dy}} \Phi(\mathbf{F}_{ex}, \mathbf{F}_{ey})+ \alpha \Psi(\mathbf{F}_{ex}, \mathbf{F}_{dx})+  \beta\Gamma(\mathbf{F}_{ey},\mathbf{F}_{dy}) 
\label{objective function}
\end{equation}
where $\Phi(\mathbf{F}_{ex}, \mathbf{F}_{ey})$ denotes the latent space loss, $\alpha\Psi(\mathbf{F}_{ex}, \mathbf{F}_{dx})$ and $\beta\Gamma(\mathbf{F}_{ey}, \mathbf{F}_{dy})$ signify the reconstruction losses. Here, $\alpha$ and $\beta$ serve to balance these components, respectively. 
In section \ref{lossfuc}, we will explain every term of the objective function in details. At the end of each epoch, we enter a validation phase, adjusting the threshold for binary label conversion by maximizing the F-measure of each label on the validation set.

After the model training is completed, we proceed with instance generation. Let $num$ represent the required number of instances to be generated, and $p$ denote the sampling rate. 
Further details on the sampling process can be found in section~\ref{generateins}.

\subsection{Loss Function \label{lossfuc}}

\subsubsection{Joint Embedding}
To calculate $\Phi(\mathbf{F}_{ex}, \mathbf{F}_{ey})$ defined in Eq.\ref{objective function}, we employ the DCCA to embed features and labels into a shared latent space simultaneously and rewrite the correlation-based $\Phi(\mathbf{F}_{ex}, \mathbf{F}_{ey})$ as the following deep version:
\begin{equation}
\Phi(\mathbf{F}_{ex}(\mathbf{X}), \mathbf{F}_{ey}(\mathbf{Y})) = \| \mathbf{F}_{ex}(\mathbf{X}) - \mathbf{F}_{ey}(\mathbf{Y}) \|_F^2 = \text{Tr}(\mathbf{C}_1^T \mathbf{C}_1) + \lambda \text{Tr}(\mathbf{C}_2^T \mathbf{C}_2 + \mathbf{C}_3^T \mathbf{C}_3)
\end{equation}
where
\begin{equation}
\begin{aligned}
\mathbf{C}_1 &= \mathbf{F}_{ex}(\mathbf{X}) - \mathbf{F}_{ey}(\mathbf{Y}), \\
\mathbf{C}_2 &= \mathbf{F}_{ex}(\mathbf{X})\mathbf{F}_{ex}(\mathbf{X})^T - \mathbf{I}, 
\\ 
\mathbf{C}_3 &= \mathbf{F}_{ey}(\mathbf{Y})\mathbf{F}_{ey}(\mathbf{Y})^T - \mathbf{I},\\
constraint: 
\mathbf{F}_{ex}&(\mathbf{X})\mathbf{F}_{ex}(\mathbf{X})^T  = \mathbf{F}_{ey}(\mathbf{Y})\mathbf{F}_{ey}(\mathbf{Y})^T = \mathbf{I}
\end{aligned}
\end{equation}
\(\mathbf{C}_1\) quantifies the discrepancy between the feature and label embeddings, while \(\mathbf{C}_2\) and \(\mathbf{C}_3\) assess how each embedded space diverges from orthonormality. The goal is to minimize these discrepancies to align the embeddings of \(\mathbf{X}\) and \(\mathbf{Y}\) closely, ensuring they remain orthonormal as dictated by the constraint. The identity matrix \(\mathbf{I} \in \mathbb{R}^{l \times l}\), serves as a benchmark for achieving this orthonormality, where \(l\) denotes the latent space dimension.
Integrating DCCA in our sampling framework not only enables a unified embedding of features and labels but also allows for their precise reconstruction from shared space through the functions \(\Psi(\mathbf{F}_{ex}(\mathbf{X}), \mathbf{F}_{dx}(\mathbf{X}))\) for features and \(\Gamma(\mathbf{F}_{ey}(\mathbf{Y}), \mathbf{F}_{dy}(\mathbf{Y}))\) for labels.

\subsubsection{Feature Reconstruction}
The function \(\Psi\) is composed of two distinct components: the feature reconstruction error, \(\mathcal{M}\), and the instance similarity metric, \(\mathcal{S}\). It is defined as:
\begin{equation}
\Psi(\mathbf{F}_{ex}(\mathbf{X}), \mathbf{F}_{dx}(\mathbf{X})) = \mathcal{M} + \lambda \mathcal{S}
\end{equation}
where \(\lambda\) is a regularization parameter that balances the contribution of the similarity metric \(\mathcal{S}\) relative to the reconstruction error \(\mathcal{M}\). 

The reconstruction error \(\mathcal{M}\), quantified as mean squared error, is calculated as:
\begin{equation}
\mathcal{M} = \sum_{i=1}^{n} (\mathbf{x}'_i - \mathbf{x}_i)^2
\end{equation}
with \(\mathbf{x}'_i\) representing the reconstruction of the input \(\mathbf{x}_i\), generated by the \(\mathbf{F}_{dx}\) applied to the encoded representation \(\mathbf{F}_{ex}(\mathbf{x}_i)\).

The similarity metric \(\mathcal{S}\) ensures that the proximity between original instances is maintained after reconstruction, thereby conserving the integrity of the feature space. This metric is formulated as:
\begin{equation}
\mathcal{S} = \frac{1}{n(n - 1)} \sum_{\substack{i,j=1,i \neq j}}^{n} \left( (\mathbf{x}_i - \mathbf{x}_j)^2 - (\mathbf{x}'_i - \mathbf{x}'_j)^2 \right)^2
\end{equation}
which measures the squared differences in distances between all pairs of original instances \((\mathbf{x}_i, \mathbf{x}_j)\) and their corresponding reconstructed pairs \((\mathbf{x}'_i, \mathbf{x}'_j)\), ensuring the model preserves instance similarity in its learned feature space. 
The combination of \(\mathcal{M}\) and \(\mathcal{S}\) ensures an optimal balance between high-fidelity feature reconstruction and the preservation of relative distances among data within the feature space.

\subsubsection{Label Reconstruction}
The function $\Gamma$ encapsulates ranking loss to help the model retrieve label vectors from the shared embedding space:
\begin{equation}
\Gamma(\mathbf{F}_{ey}(\mathbf{Y}), \mathbf{F}_{dy}(\mathbf{Y})) = \sum_{i=1}^{n} \left( \frac{E_i}{|Y_i| \times |\bar{Y_i}|} \right)
\label{labelloss}
\end{equation}
where $E_i$ is defined as the set of label pairs $(y_{ij}, y_{ik})$ that satisfy the condition $f(\mathbf{x}_i, y_{ij}) \leq f(\mathbf{x}_i, y'_{ij})$, with these label pairs belonging to the Cartesian product of the set of positive labels $Y_i$ and the set of negative labels $\bar{Y_i}$. Here, $Y_i$ represents the set of positive labels, while $\bar{Y_i}$ represents the set of negative labels.

\subsection{Generate Instances and Post-Processing\label{generateins}}
Let $L_{s}= \left \{ l_j \mid ImR_j > 10 ,IRlbl_j > MeanIR \right \}$ \footnote{Here, 10 is a hyperparameter. We refer to the suggestions in \cite{MLBOTE} for the selection.} be the set comprising $m$ minority labels \cite{MLBOTE} and $M=\left \{ \left ( \mathbf{x}_i,\mathbf{y}_i \right)\mid y_{ij}=1, l_j \in L_s\right \}$ be the minority instance set associated any labels in $L_{s}$. 
Then, we randomly select a seed sample $(\mathbf{x}_i,\mathbf{y}_i)$ from $M$ to initiate the sampling process through forward inference.
\begin{figure}[!h]
\centering
\includegraphics[width=0.8\textwidth]{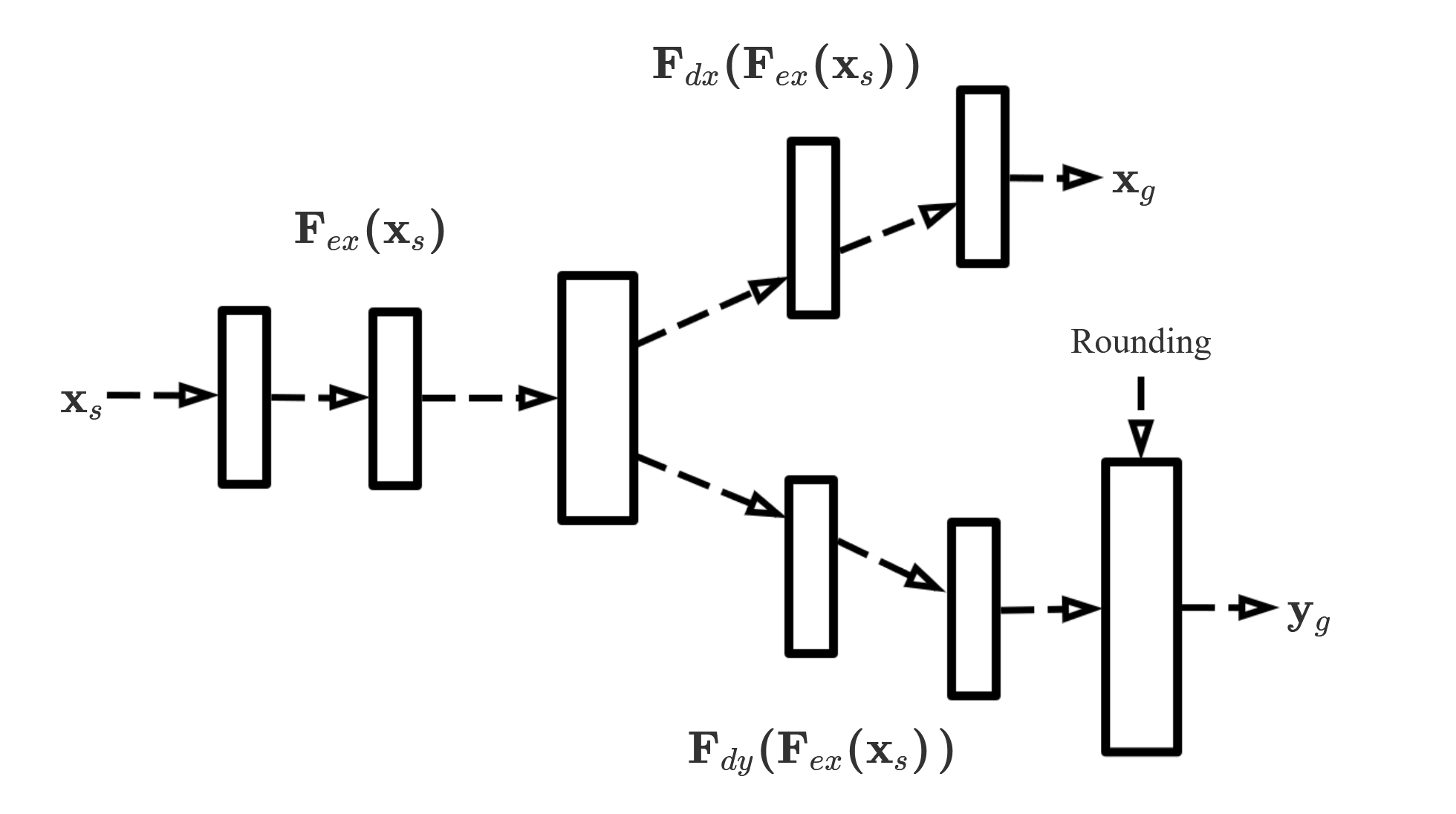} 
\caption{The process of generating instances}
\label{fig:sampling}
\end{figure}
As shown in Figure \ref{fig:sampling}, the process encodes the feature vector $\mathbf{x}_s$ into a latent space by $\mathbf{F}_{ex}(\mathbf{x}_s)$, then decodes it to the feature and label vectors of the new instance by $\mathbf{F}_{dx}(\mathbf{F}_{ex}(\mathbf{x}_s))$ and $\mathbf{F}_{dy}(\mathbf{F}_{ex}(\mathbf{x}_s))$, respectively. Specifically, we employ a predefined threshold set $T$ to transform the decoded numerical label vector into a binary label vector. 
After the process, we remove any instances where the generated label vector is entirely zeros to ensure each instance contributes meaningfully to the dataset.

\section{Experiments and Analysis}
\subsection{Datasets}
We evaluate our proposed model across 9 benchmark multi-label datasets spanning diverse domains, such as text, images, and bioinformatics \cite{MULAN}. Each dataset is characterized by a set of statistics and imbalance metrics, which include the number of instances (\(n\)), feature dimensions (\(d\)), labels (\(q\)), average label cardinality per instance \(Card(q)\), and label density \(Den(q)\). A comprehensive explanation of these statistical measures and imbalance metrics is available in \cite{mulitilabelimbalancereview,mulitilabellearningreview}. In the experiment, 20\% training data is split as the validation set, which is used to establish thresholds for accurate prediction of final labels.

\begin{table}[!ht]
\centering
\caption{Characteristics of the experimental datasets}
\label{ta:mld}
\resizebox{0.9\textwidth}{!}{
\renewcommand{\arraystretch}{1}
\begin{tabular}{cccccccccccc}
\toprule
Dataset & Domain & $n$ & $d$ & $q$ & $Card(q)$ & $Den(q)$ & MeanIR & CVIR \\  \midrule
bibtex & text & 7395 & 1836 & 159 & 2.40 & 0.02 & 12.50 & 0.41 \\
enron & text & 1702 & 1001 & 53 & 3.38 & 0.06 & 73.95 & 1.96 \\
Languagelog& text & 1460 & 1004 & 75 & 15.93 & 0.21 & 5.39 & 0.78 \\
yeast & biology & 2417 & 103 & 14 & 4.24 & 0.30 & 7.20 & 1.88  \\
rcv1 & text & 6000 & 472 & 101 & 2.88 & 0.03 & 54.49 & 2.08  \\
rcv2 & text & 6000 & 472 & 101 & 2.63 & 0.03 & 45.51 & 1.71  \\
rcv3 & text &  6000 & 472 & 101 & 2.61 & 0.03 & 68.33 & 1.58 \\
cal500 & music & 502 & 68 & 174 & 26.04 & 0.15 & 20.58 & 1.09  \\
Corel5k & images & 5000 & 499 & 374 & 3.52 & 0.01 & 189.57 & 1.53 \\

\bottomrule
\end{tabular}
}
\end{table}


\subsection{Experiment Setup}
In AEMLO, \(\mathbf{F}_{ex}\) and \(\mathbf{F}_{ey}\) are comprised of two fully connected layers, whereas \(\mathbf{F}_{dx}\) and \(\mathbf{F}_{dy}\) adopt a single fully connected layer structure.   
Each layer within these components incorporates 512 neurons and incorporates a leaky ReLU activation function to introduce nonlinearity. 
The parameters \(\alpha\) and \(\beta\) of objective function are explored within the range of \([2^{-4},2^{-3},\cdots, 2^{4}]\). 

\subsubsection{Compared Sampling Methods}
We compare our proposed sampling method with the following sampling methods. 
\begin{itemize}
    \item \textbf{MLSMOTE}: MLSMOTE extends the classical SMOTE method to multi-label data.$\left [Parameter:Ranking,k=5\right ]$
    \item \textbf{MLSOL}: MLSOL  considers local label imbalance and employs weight vectors and type matrices for seed instance selection and synthetic instance generation. $\left [Parameter: p\in \left ( 0.1,0.3,0.5,0.7,0.9\right), k=5\right ]$ 
    \item \textbf{MLROS}: MLROS executes replicating instances associated with minority labels. $\left [Parameter: p\in \left ( 0.1,0.3,0.5,0.7,0.9\right)\right ]$
    \item \textbf{MLRUS}: MLRUS executes removing instances associated with majority labels. $\left [Parameter: p\in \left ( 0.1,0.2,0.3\right)\right ]$
    \item \textbf{MLTL}: MLTL identifies and removes Tomek-Links in multi-label data by considering the set of instances associated with each minority label. $\left [Parameter:k=5 \right ]$ 
    \item \textbf{MLBOTE}: MLBOTE divides instances into three
     categories, and determines specific instance weights and sampling rates for each group.
\end{itemize}
\subsubsection{Base Multi-Lable Classifiers}
We use all sampling methods on the following five multi-label classifiers.
\begin{itemize}
    \item \textbf{Binary Relevance} \cite{BR}: BR transforms the multi-label classification problem into multiple independent binary classification tasks, each of which corresponds to one label and trains a binary classifier. Base binary classifier: SVM. 
    \item \textbf{Multi-Label k-Nearest Neighbors} \cite{MLKNN}: MLkNN is an extension of the $k$-Nearest Neighbors ($k$NNs) algorithm for multi-label classification. hyperparameter configuration: $k$=10.
    \item \textbf{Random k-labELsets} \cite{RAkEL}: RAkEL divides the entire label set into several random subsets containing at least three labels and encodes each subset as a multi-class dataset by treating each label combination as a class. hyperparameter configuration: k=3, n=2q, base binary classifier: C4.5 Decision Tree.
    \item \textbf{Ensemble of Classifier Chain} \cite{ECC}: ECC is an approach that extends the Classifier Chain further in an ensemble framework. hyperparameter configuration: N = 5, base binary classifier: C4.5 Decision Tree.
    \item \textbf{Calibrated Label Ranking} \cite{CLR}: CLR transforms the multi-label learning problem into the label ranking problem. Base binary classifier: SVM
\end{itemize}
\subsubsection{Evaluation Metrics}
To assess the efficacy of the batch method in multi-label classification, three commonly utilized evaluation metrics are adopted, comprising Macro-F, Macro-AUC, Ranking Loss. Please refer to \cite{mulitilabellearningreview} for detailed definitions of these metrics.

\subsection{Experimental Analysis}
\begin{figure}[!ht]
\centering
\subfigure[bibtex]{\includegraphics[width=0.3\textwidth]{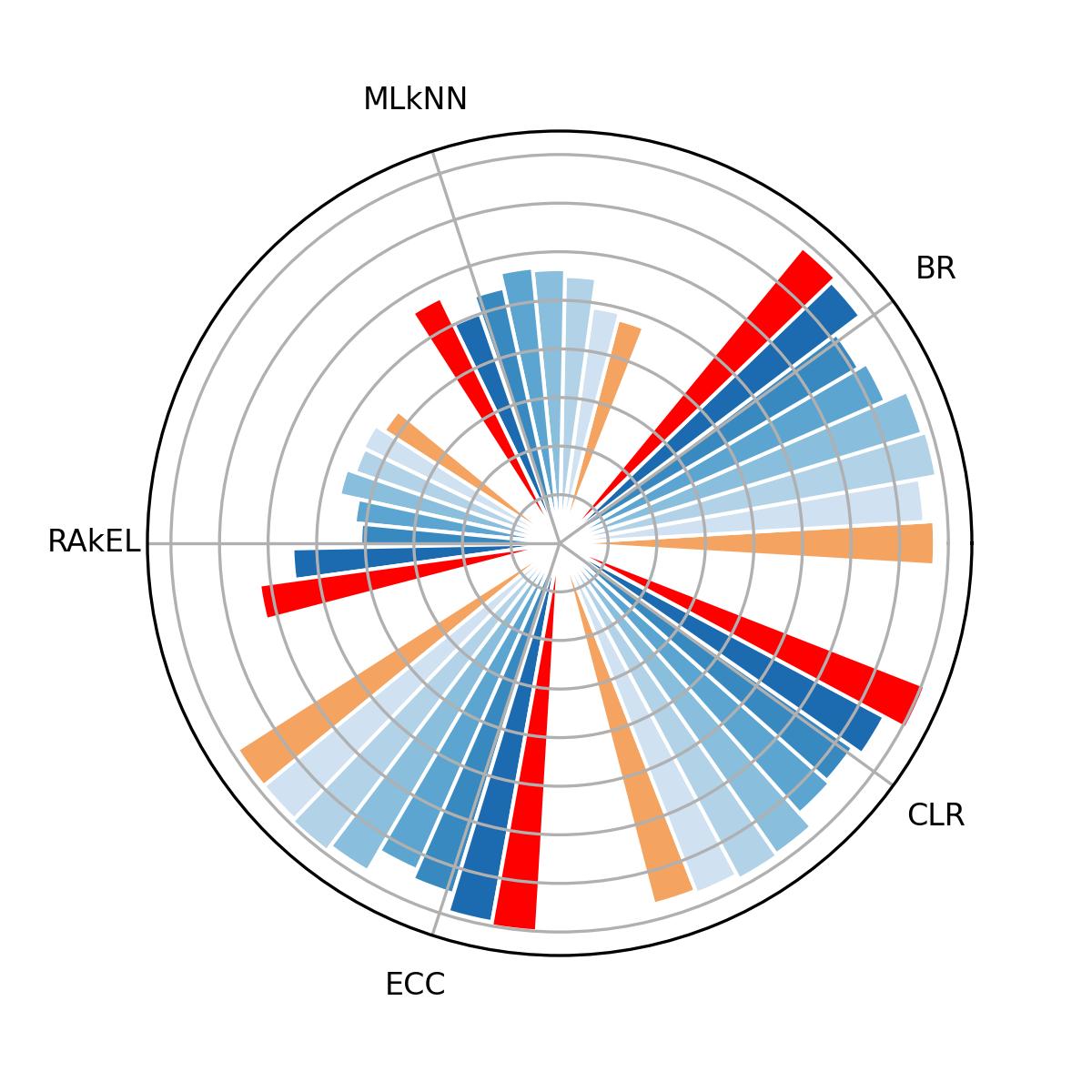}}
\subfigure[enron]{\includegraphics[width=0.3\textwidth]{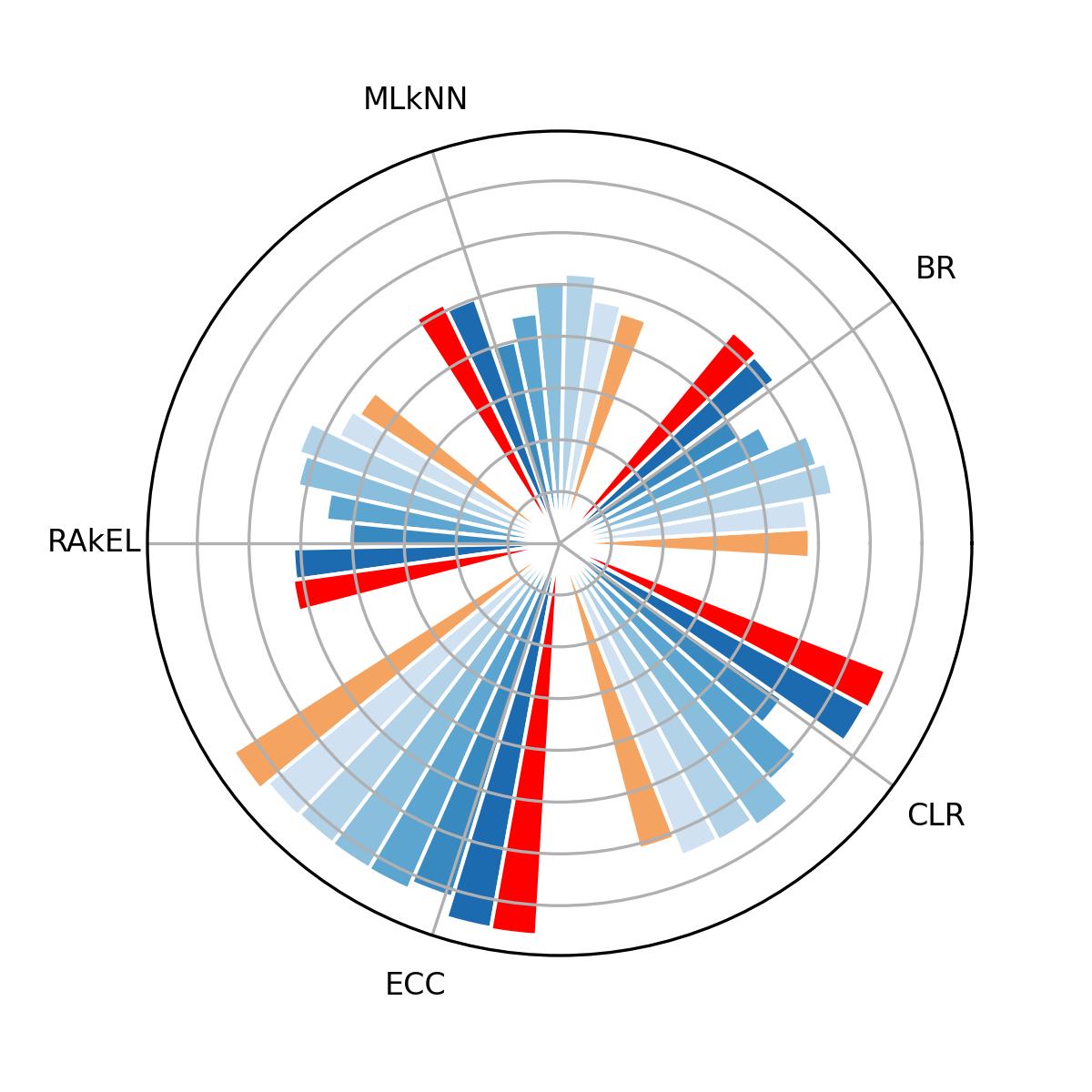}}
\subfigure[Languagelog]{\includegraphics[width=0.3\textwidth]{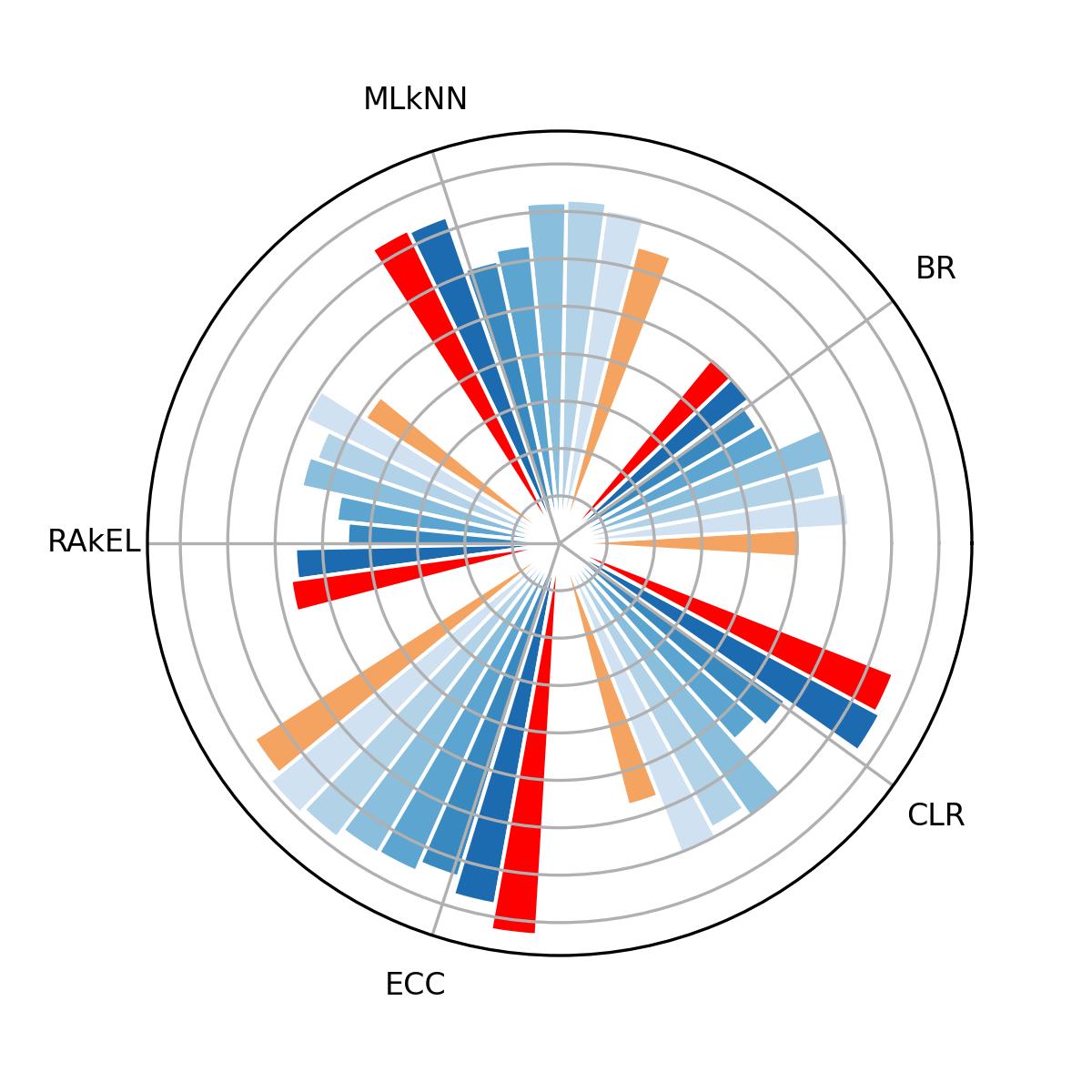}}
\subfigure[yeast]{\includegraphics[width=0.3\textwidth]{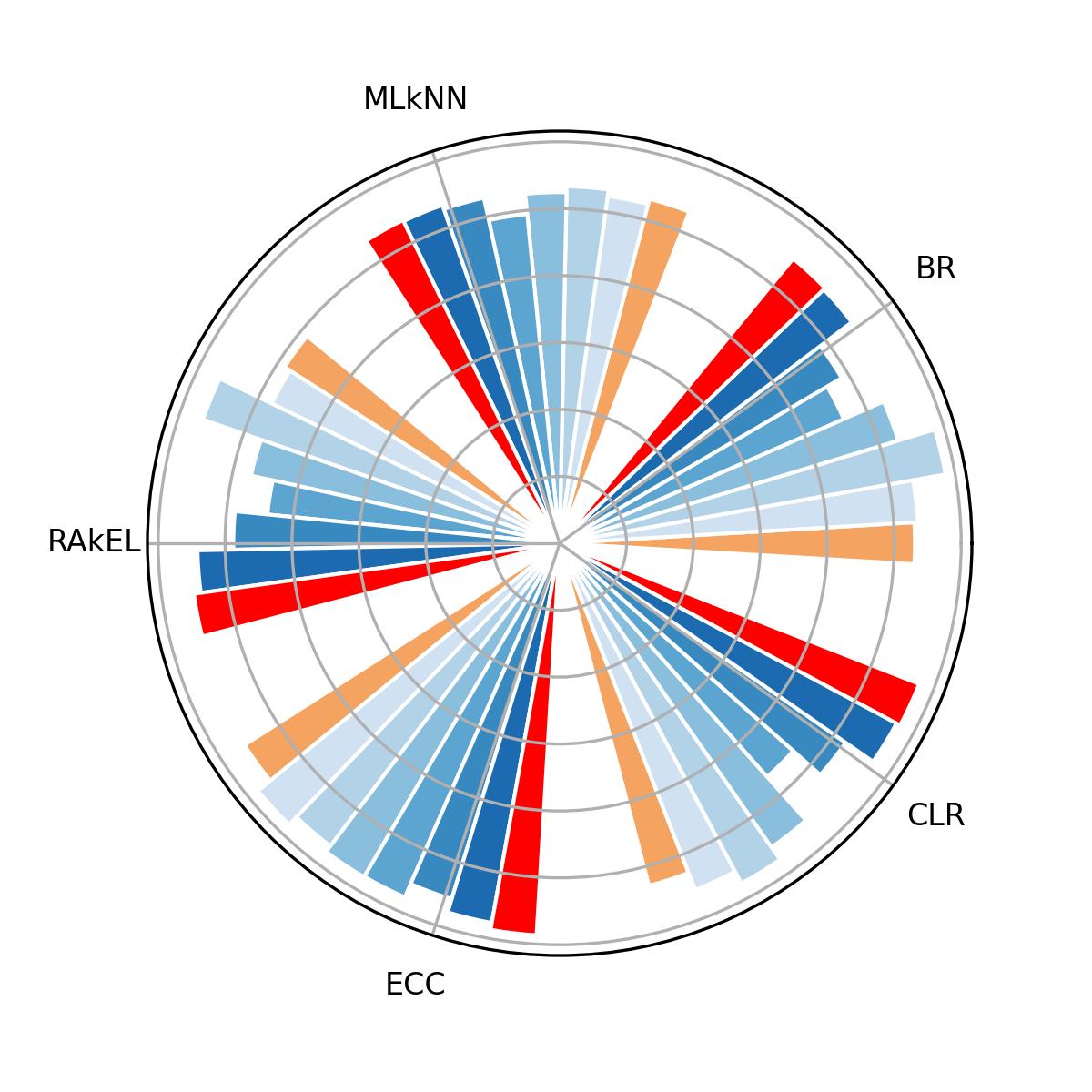}}
\subfigure[rcv1]{\includegraphics[width=0.3\textwidth]{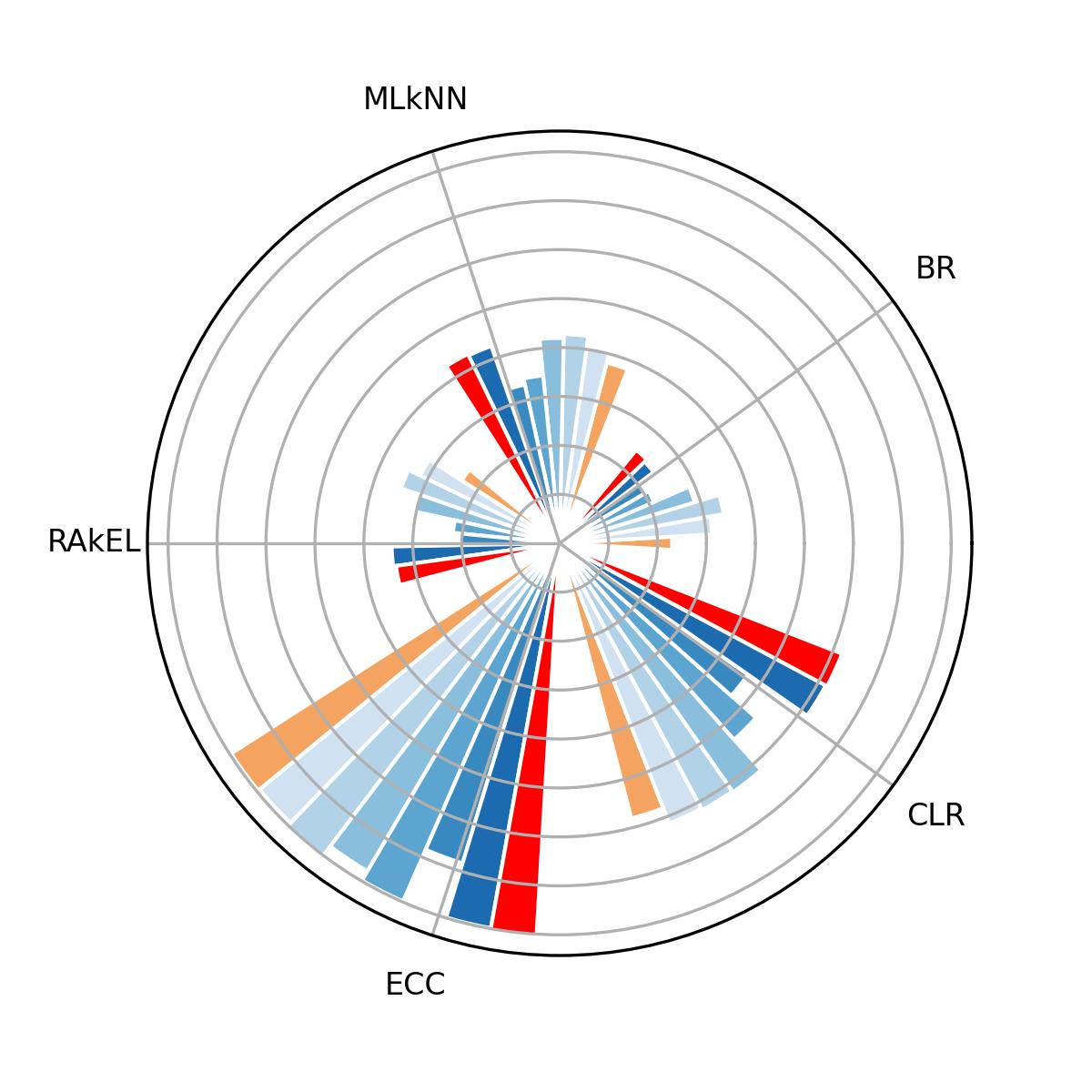}} 
\subfigure[rcv2]{\includegraphics[width=0.3\textwidth]{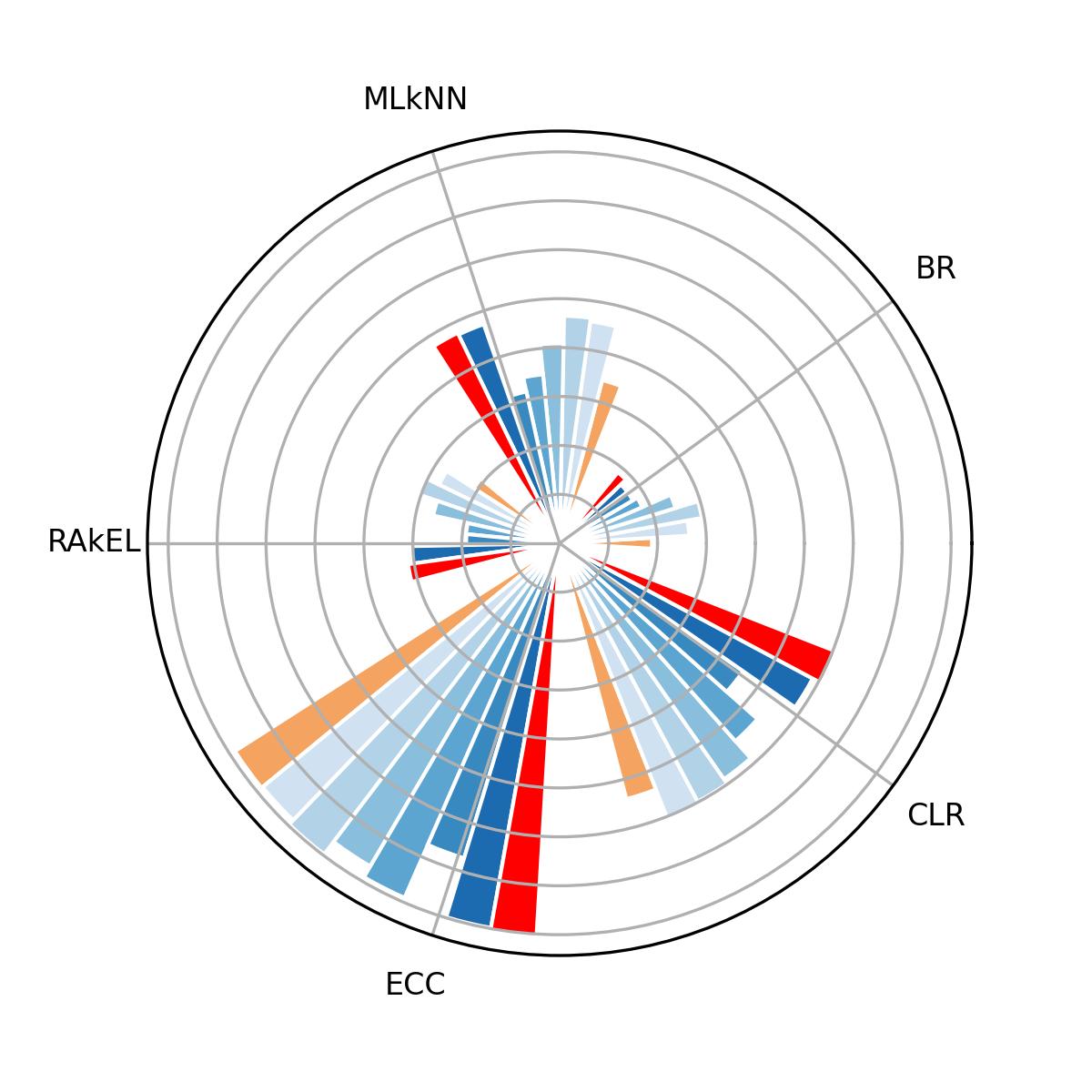}} 
\subfigure[rcv3]{\includegraphics[width=0.3\textwidth]{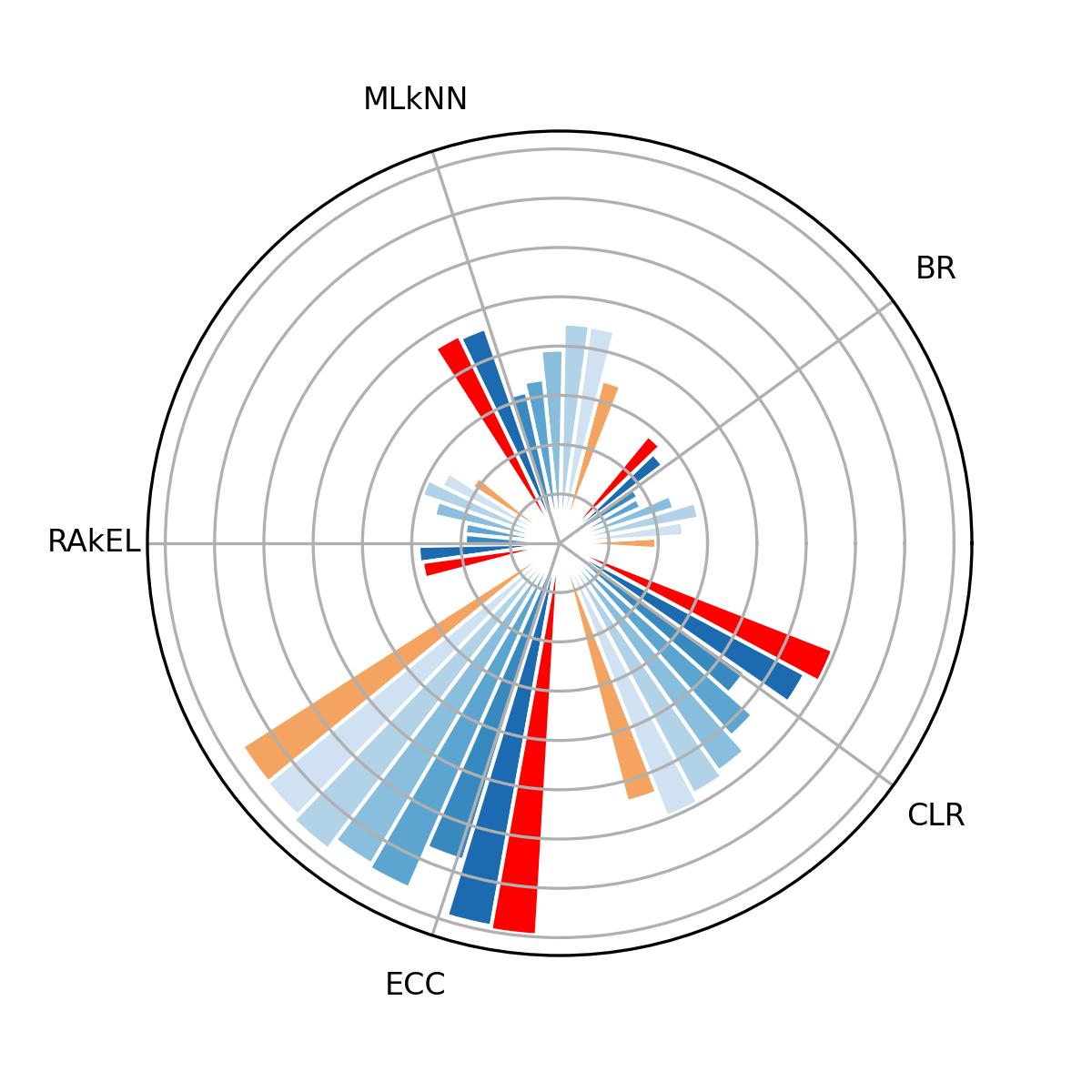}}
\subfigure[cal500]{\includegraphics[width=0.3\textwidth]{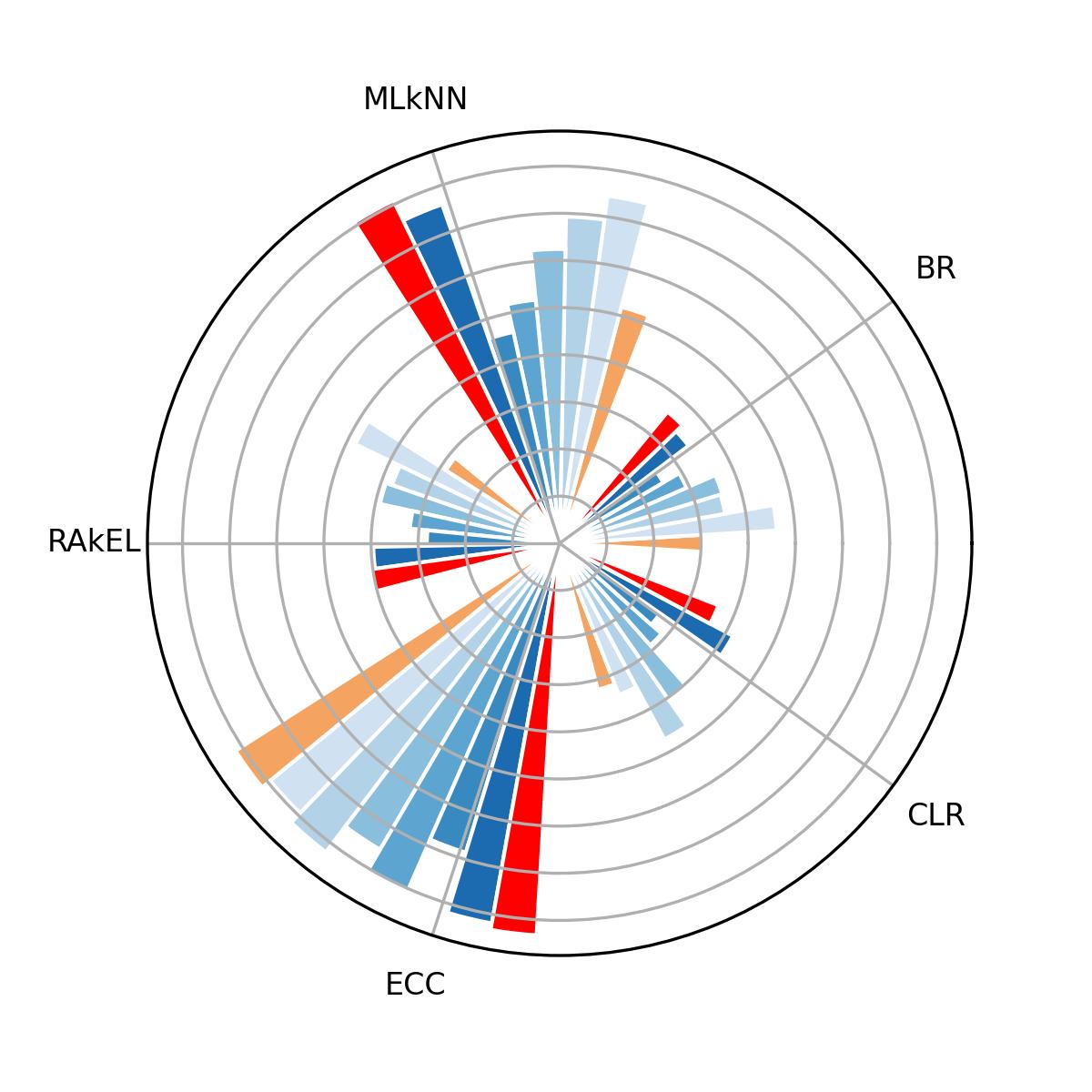}}
\subfigure[Corel5k]{\includegraphics[width=0.3\textwidth]{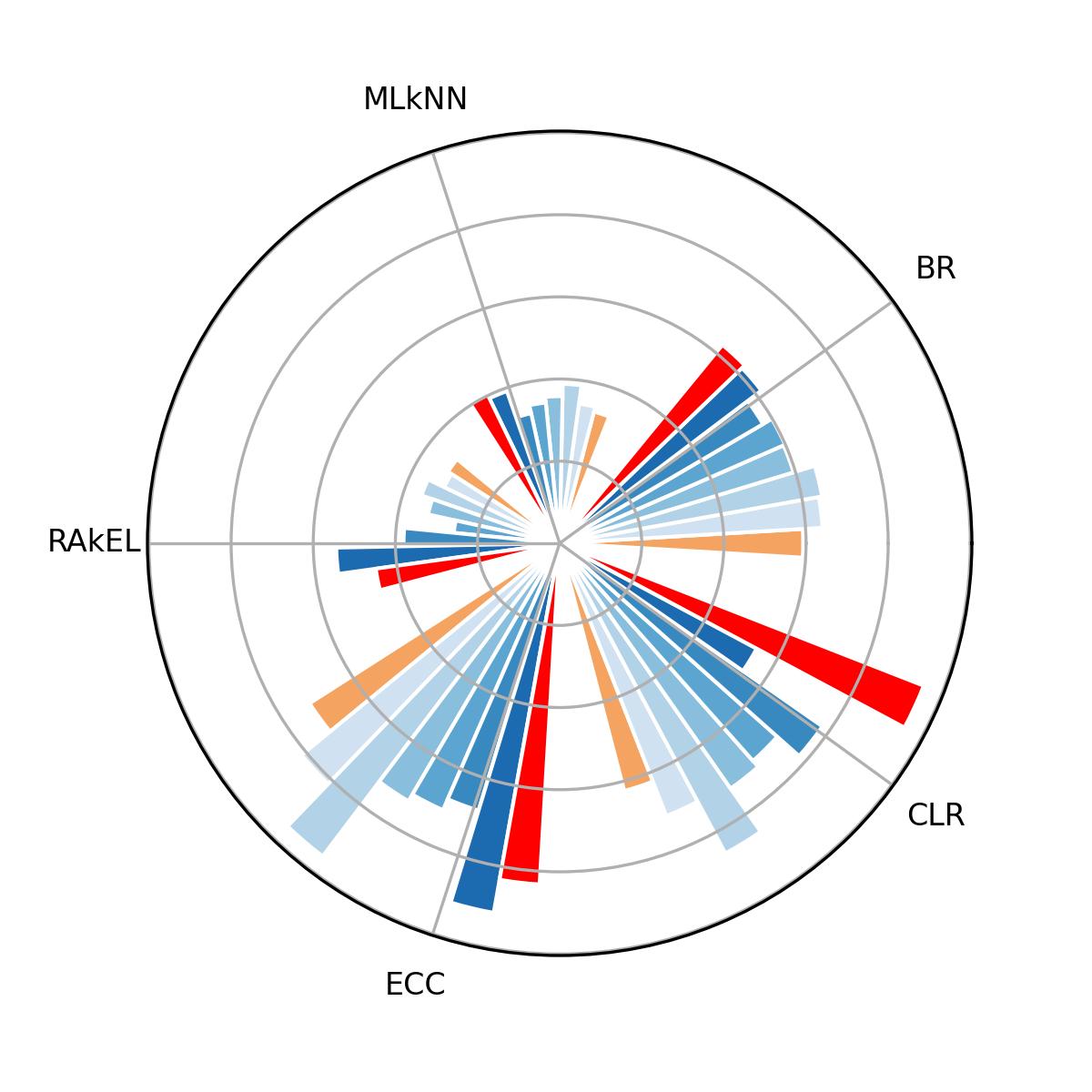}}
\subfigure{\includegraphics[width=0.9\textwidth]{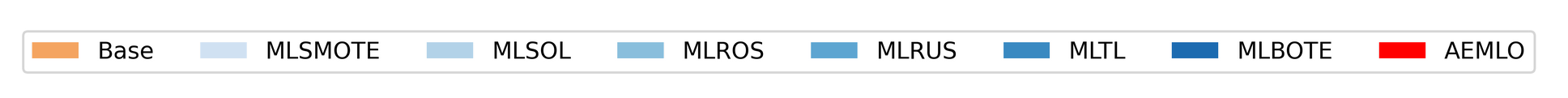}}
\caption{The performance of the multi-label sampling methods in terms of Macro-F across five different classification methods.}
\label{fig:Macro-F}
\end{figure}

\begin{figure}[!ht]
\centering
\subfigure[bibtex]{\includegraphics[width=0.3\textwidth]{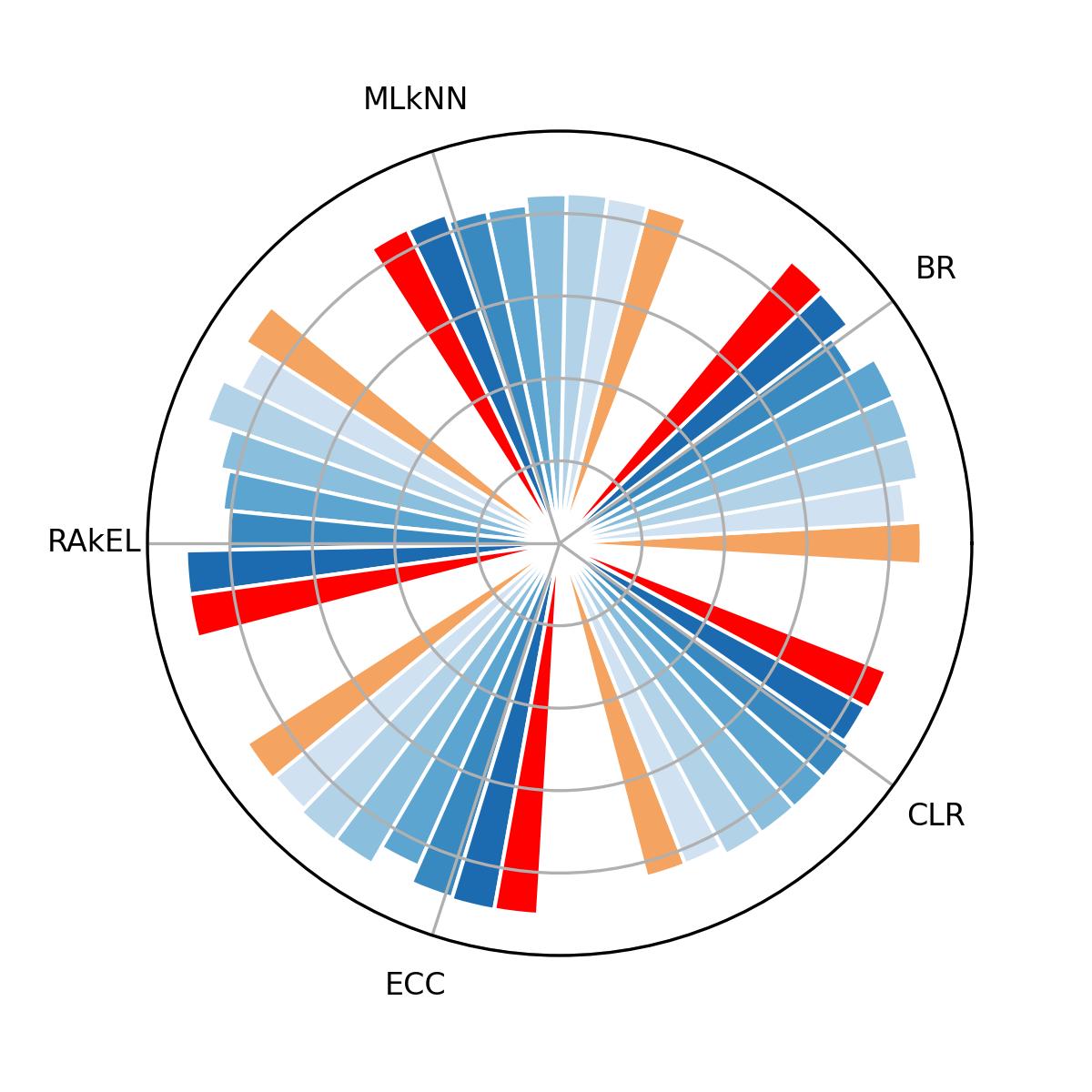}}
\subfigure[enron]{\includegraphics[width=0.3\textwidth]{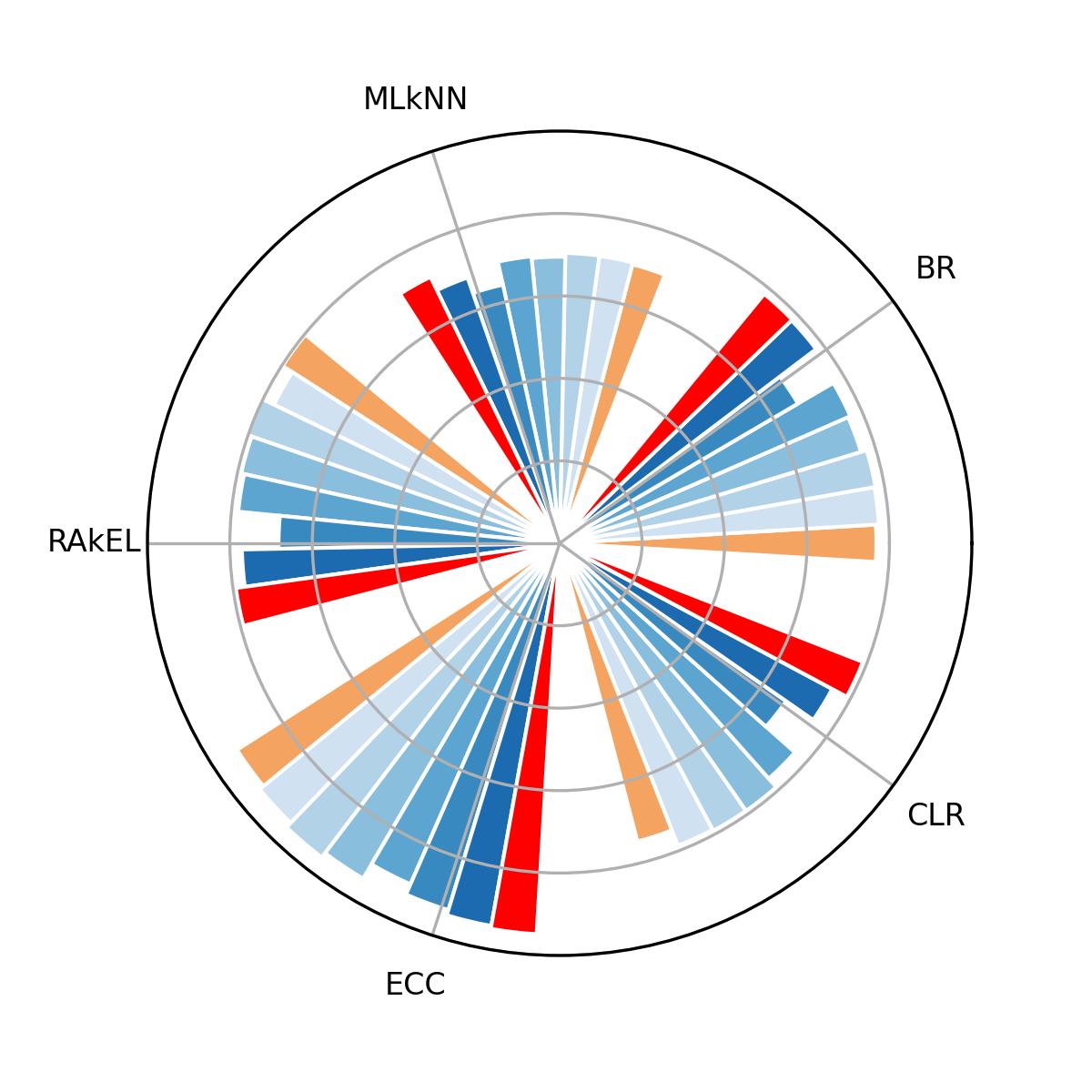}}
\subfigure[Languagelog]{\includegraphics[width=0.3\textwidth]{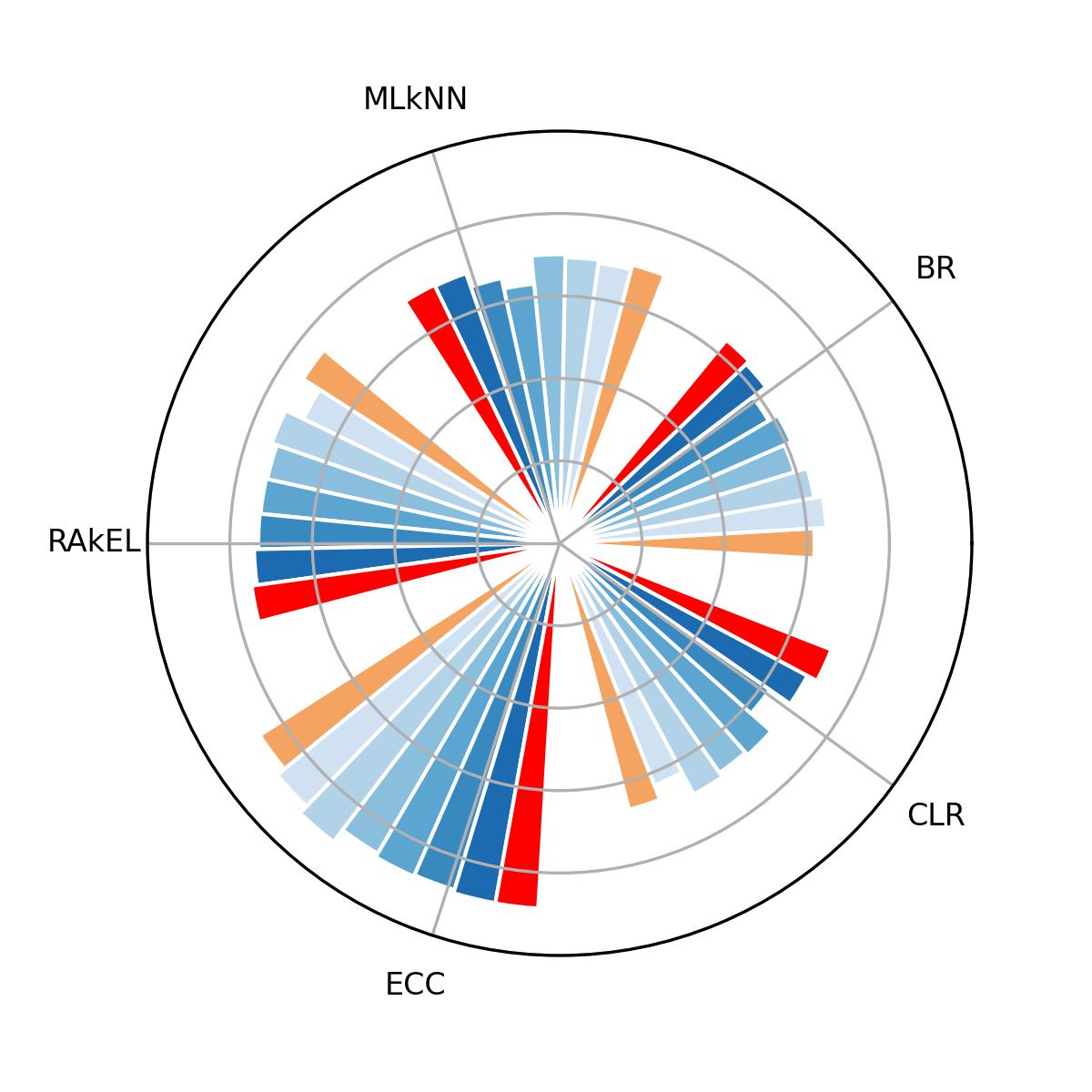}}
\subfigure[yeast]{\includegraphics[width=0.3\textwidth]{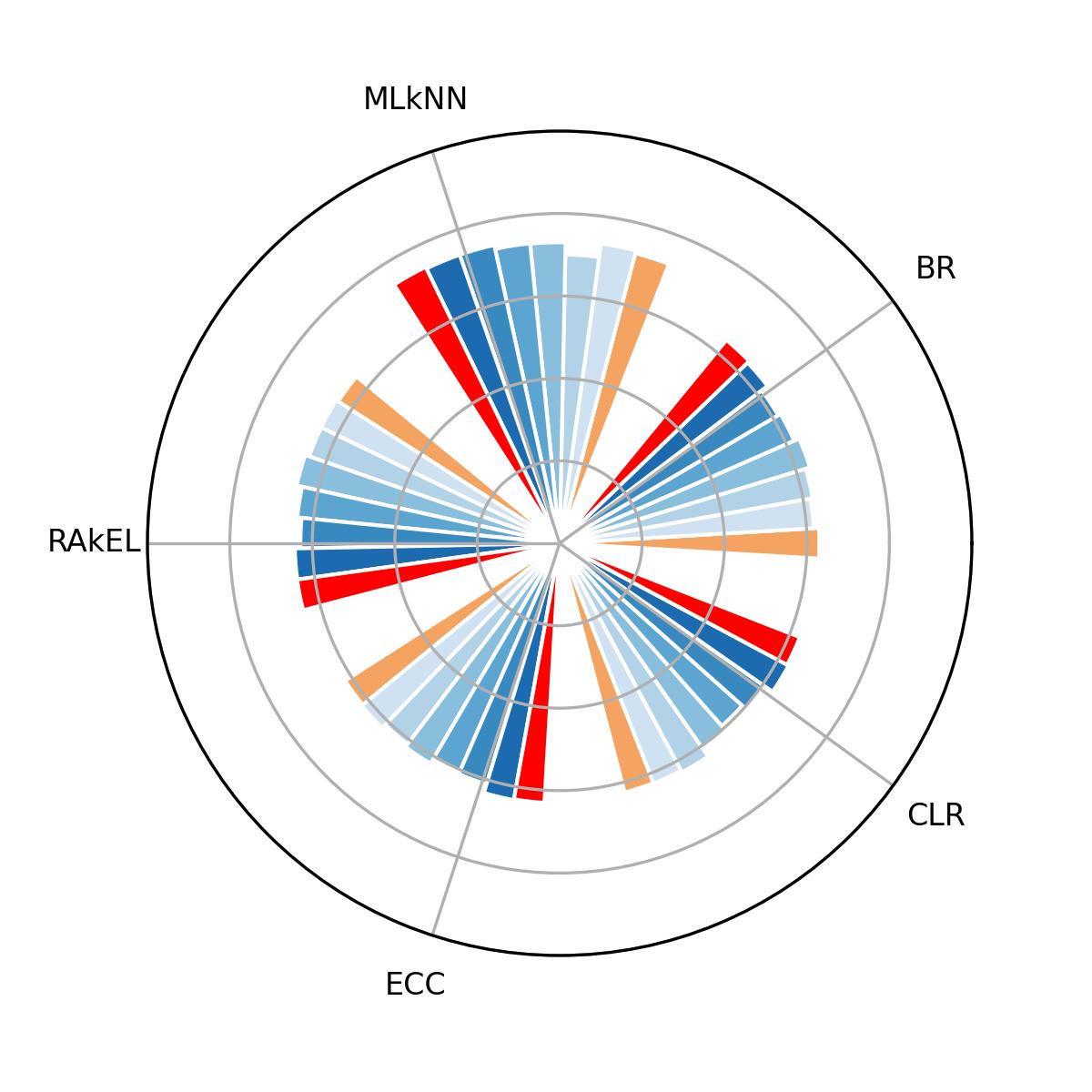}}
\subfigure[rcv1]{\includegraphics[width=0.3\textwidth]{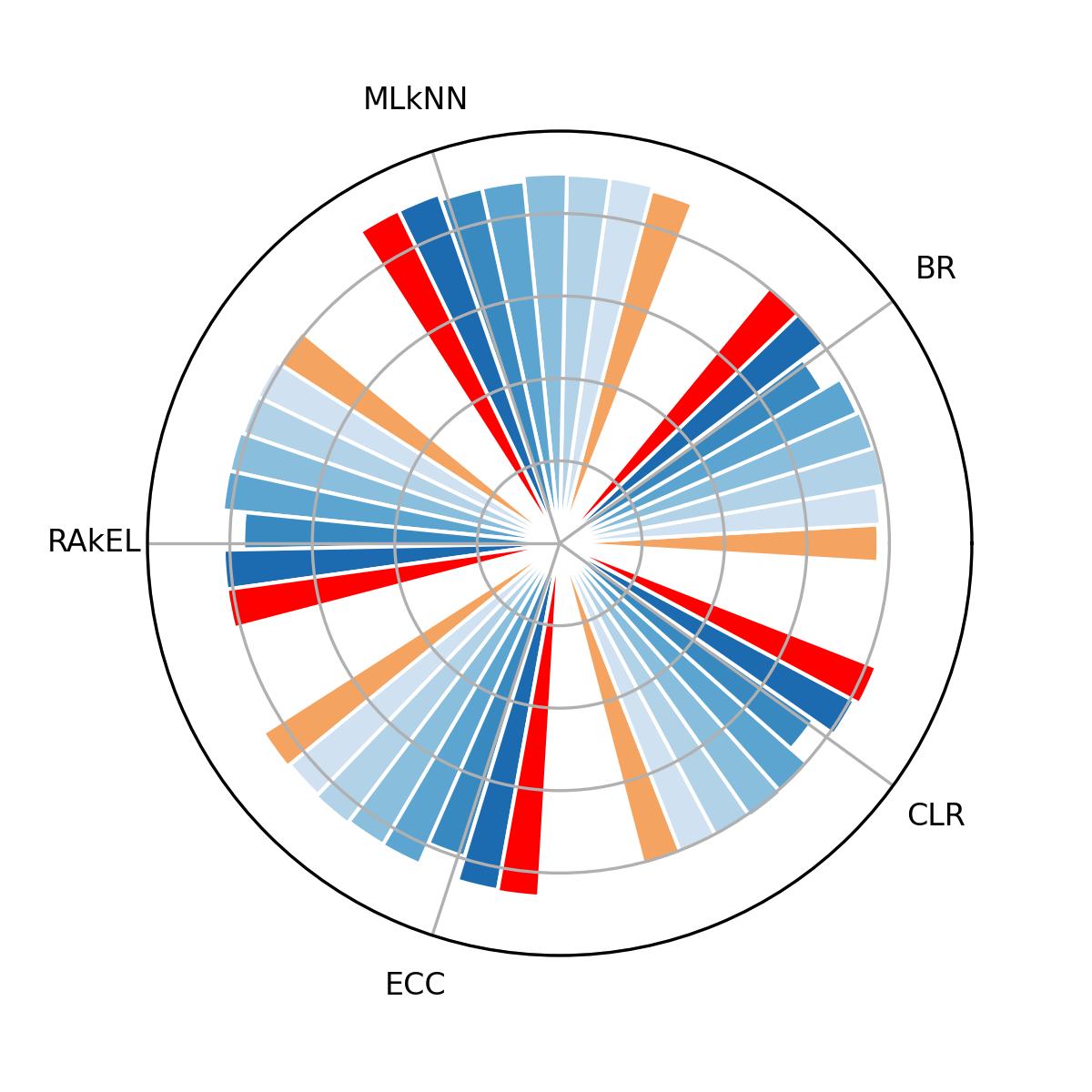}} 
\subfigure[rcv2]{\includegraphics[width=0.3\textwidth]{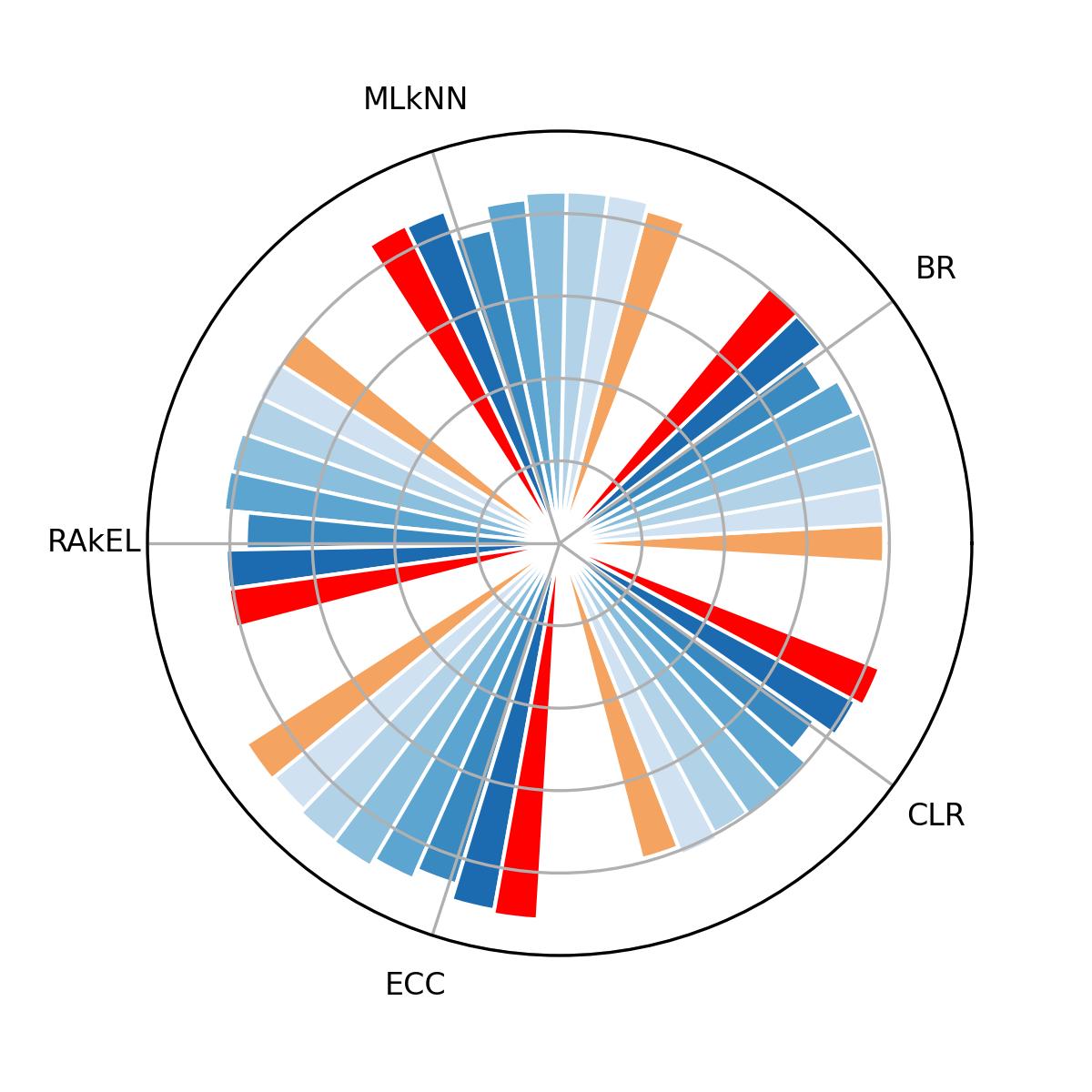}} 
\subfigure[rcv3]{\includegraphics[width=0.3\textwidth]{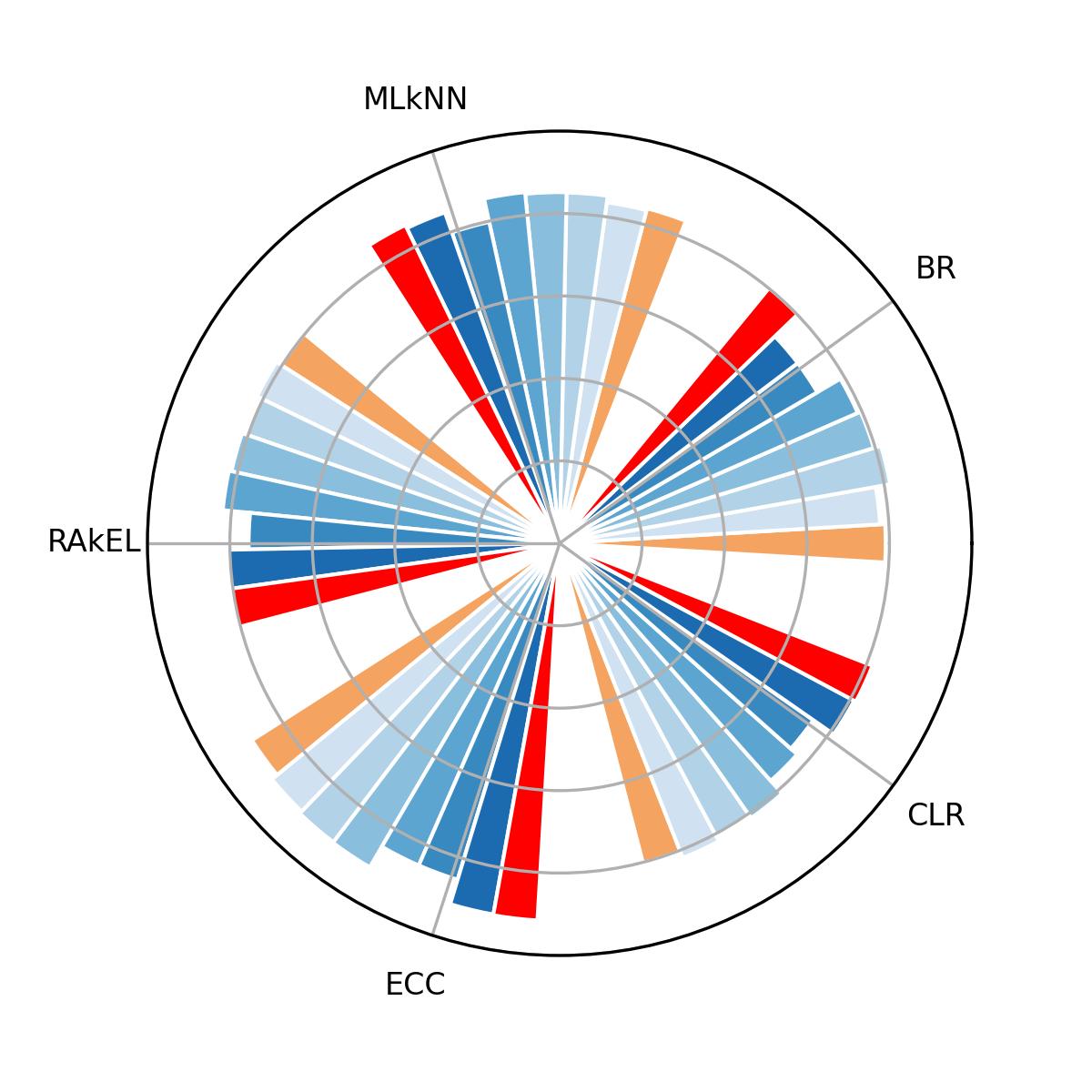}}
\subfigure[cal500]{\includegraphics[width=0.3\textwidth]{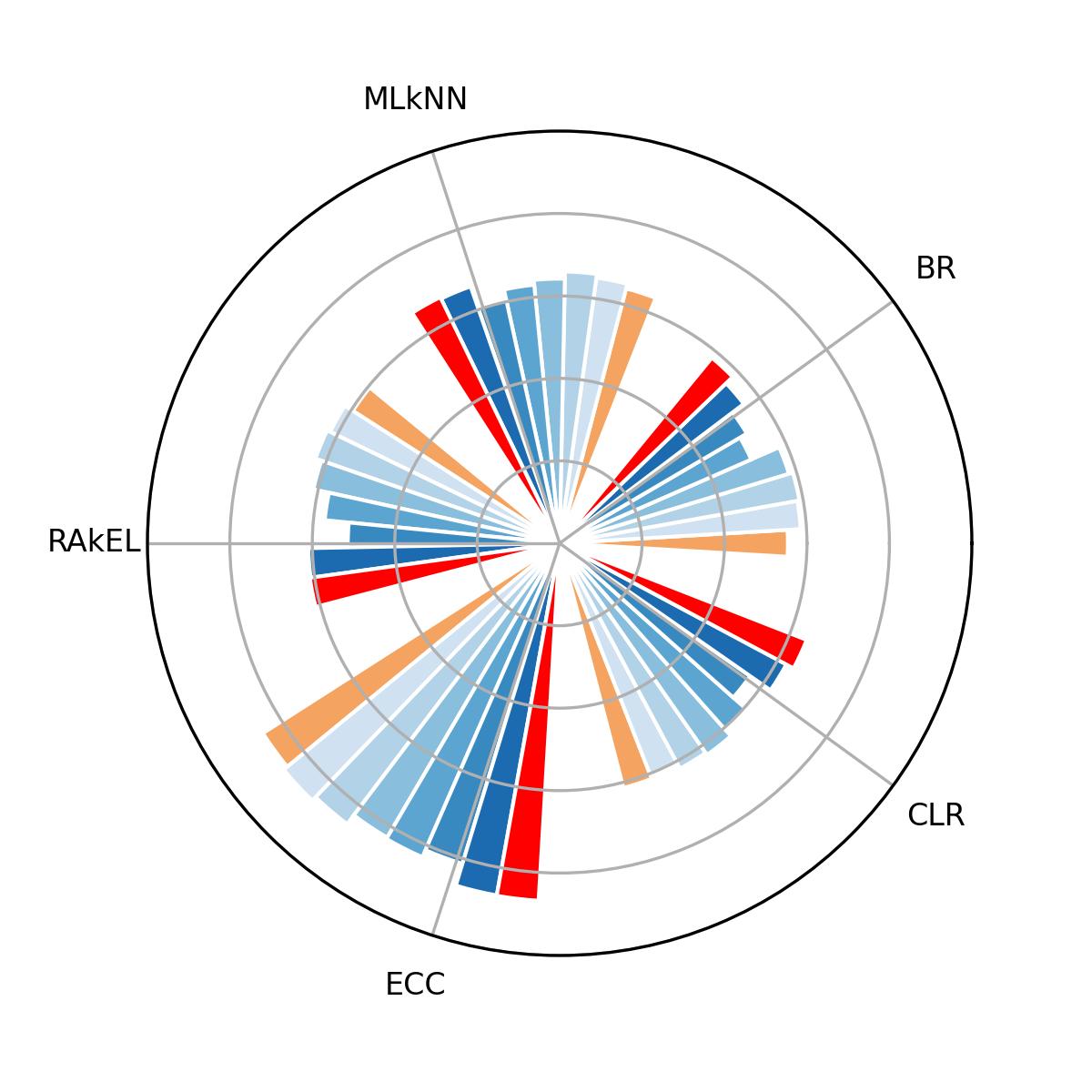}}
\subfigure[Corel5k]{\includegraphics[width=0.3\textwidth]{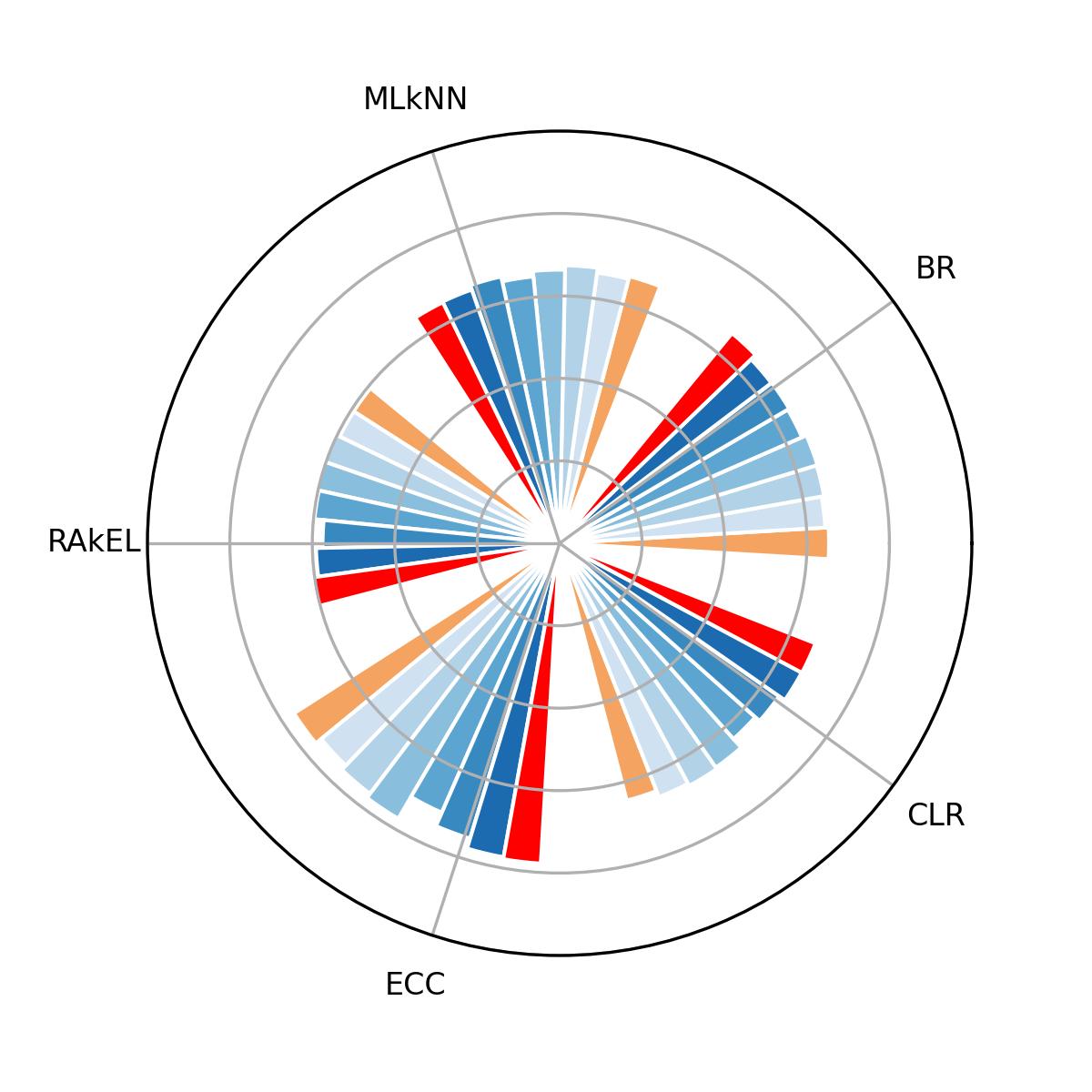}}
\subfigure{\includegraphics[width=0.9\textwidth]{legend.png}}
\caption{The performance of the multi-label sampling methods in terms of Macro-AUC across five different classification methods.}
\label{fig:Macro-AUC}
\end{figure}

\begin{figure}[!ht]
\centering
\subfigure[bibtex]{\includegraphics[width=0.28\textwidth]{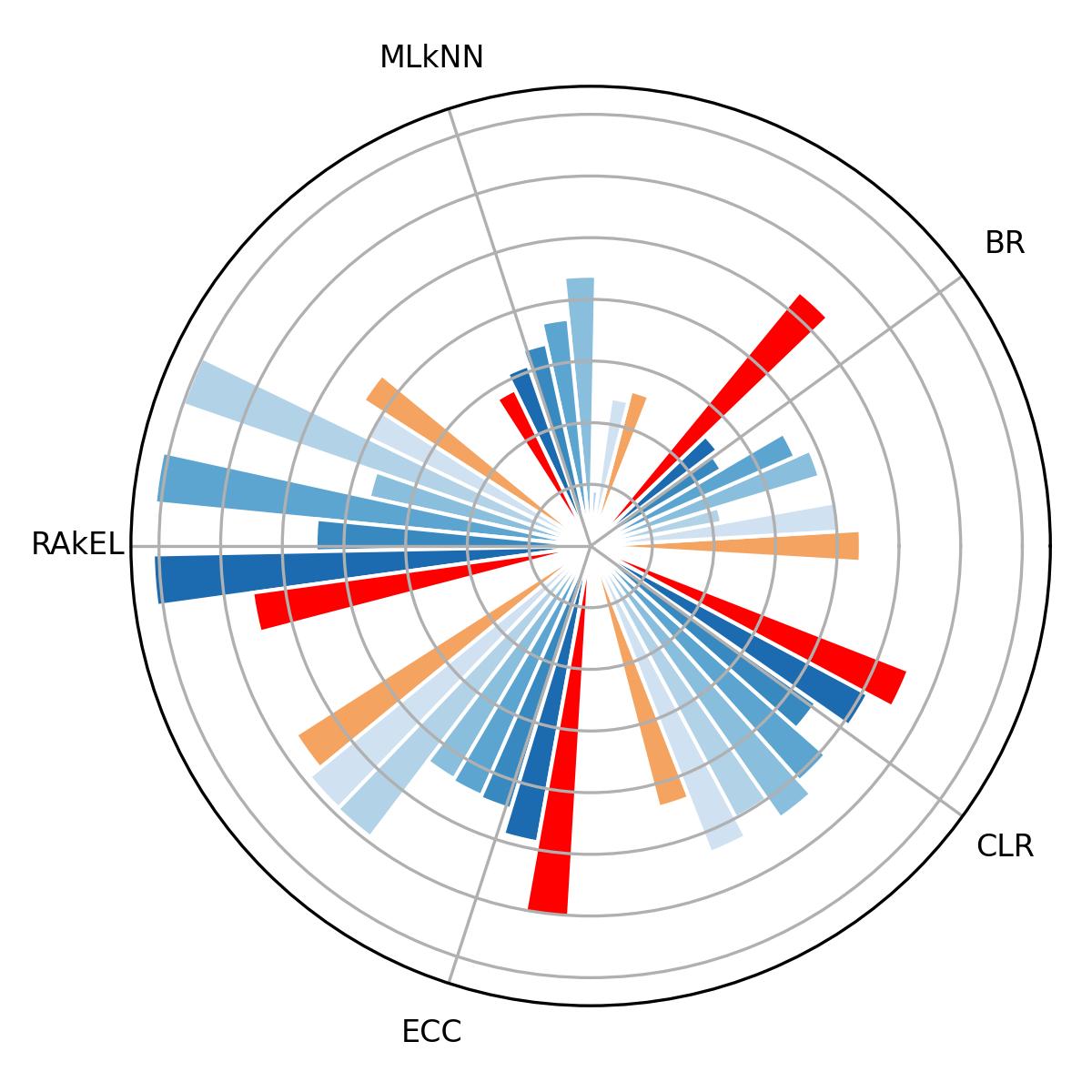}}
\subfigure[enron]{\includegraphics[width=0.28\textwidth]{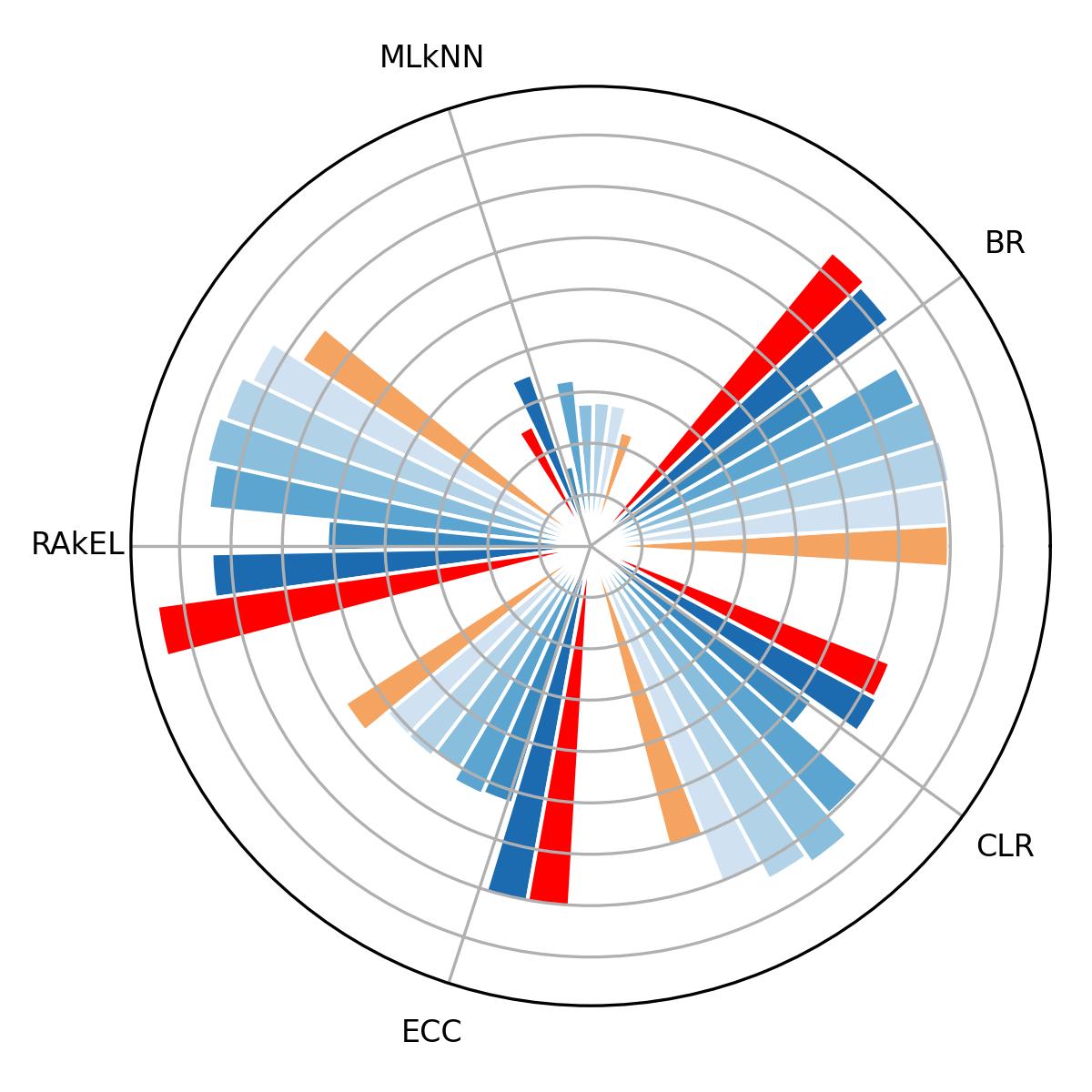}}
\subfigure[Languagelog]{\includegraphics[width=0.28\textwidth]{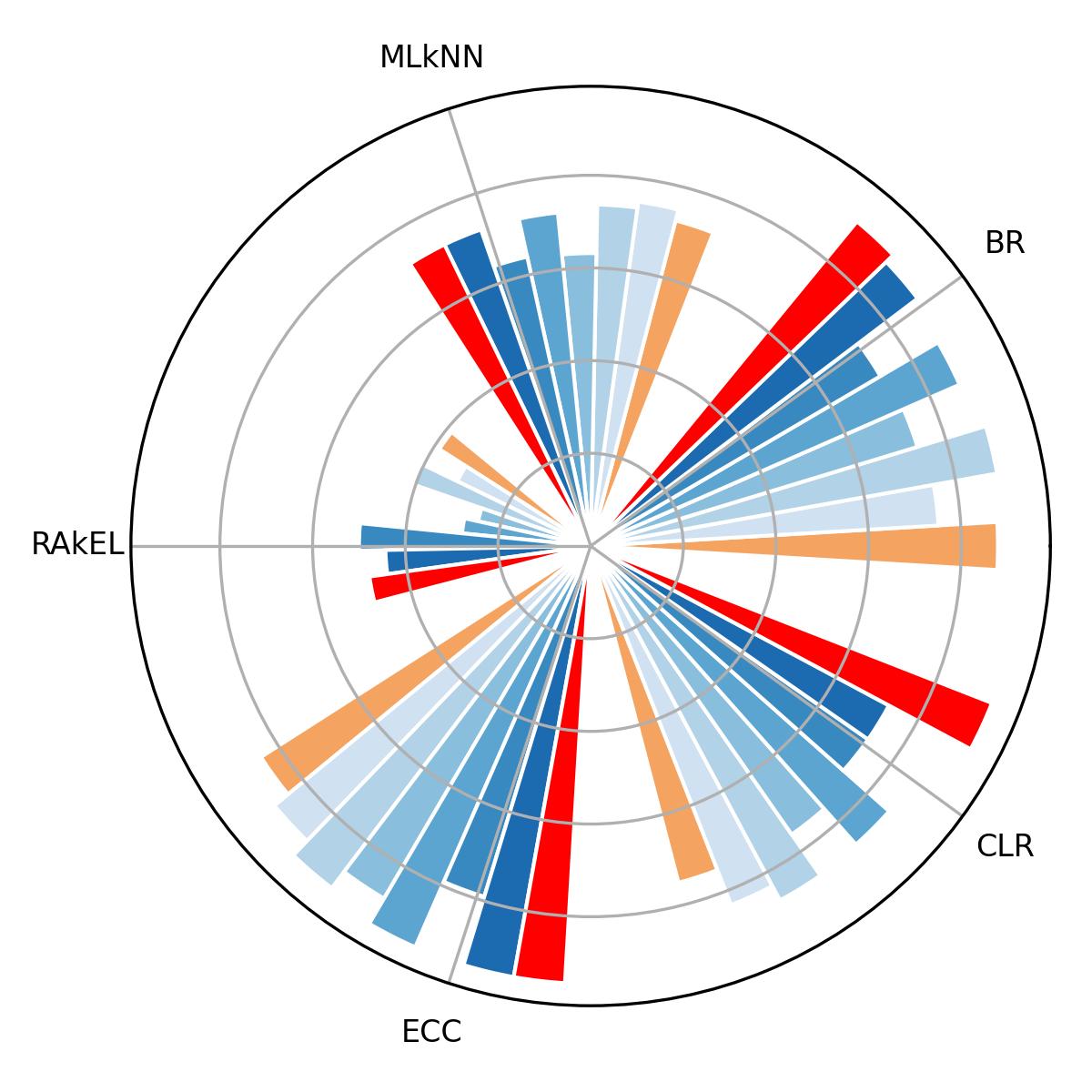}}
\subfigure[yeast]{\includegraphics[width=0.28\textwidth]{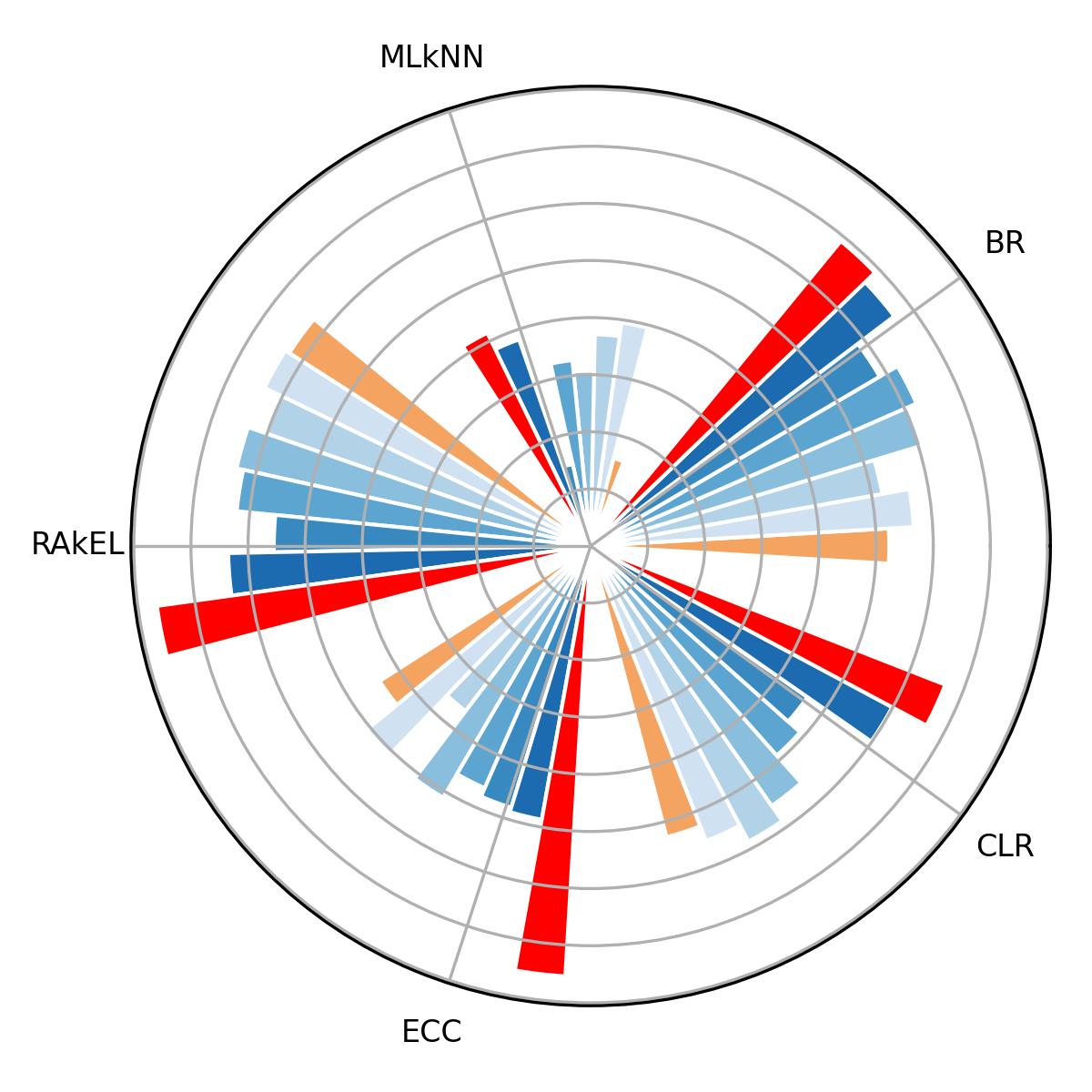}}
\subfigure[rcv1]{\includegraphics[width=0.28\textwidth]{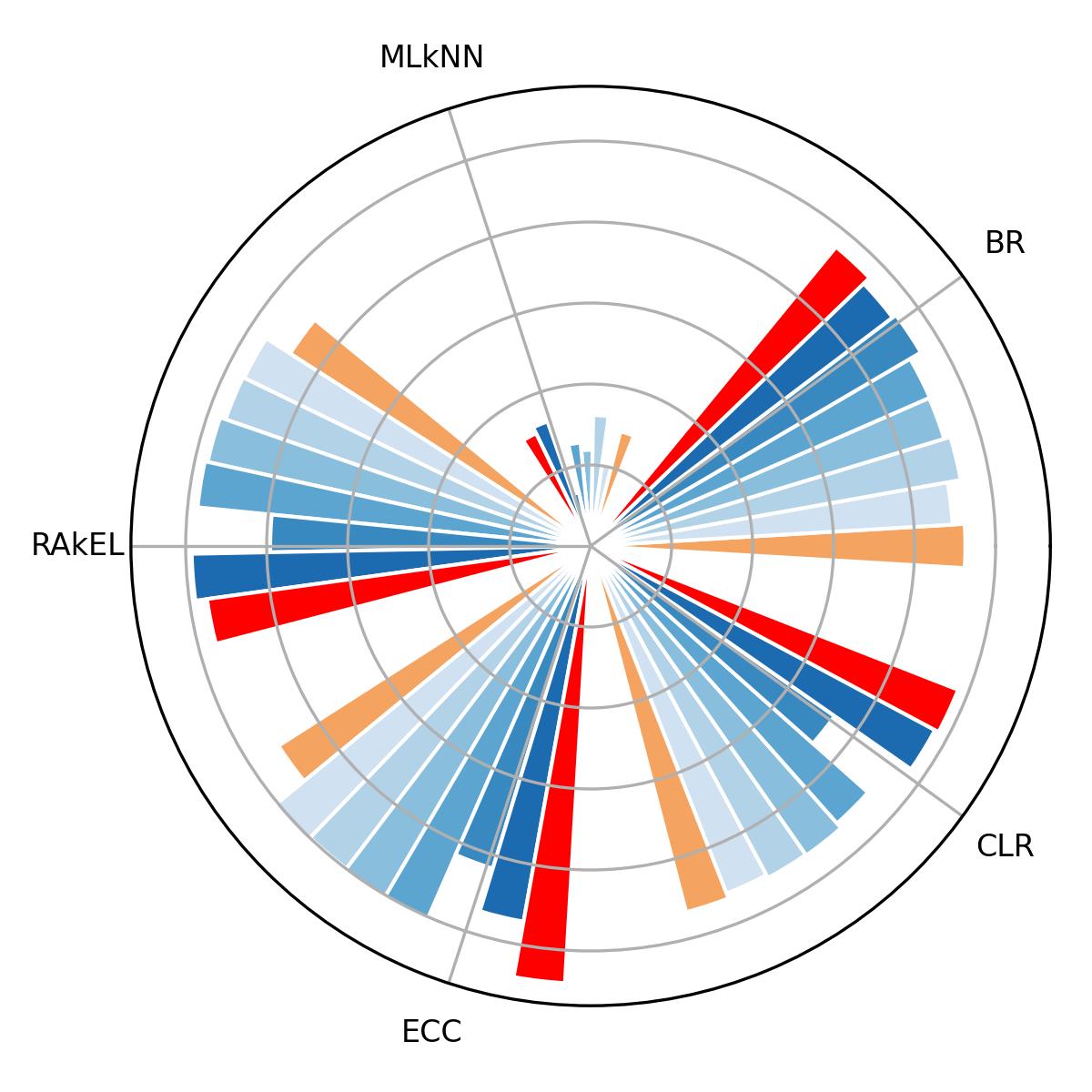}} 
\subfigure[rcv2]{\includegraphics[width=0.28\textwidth]{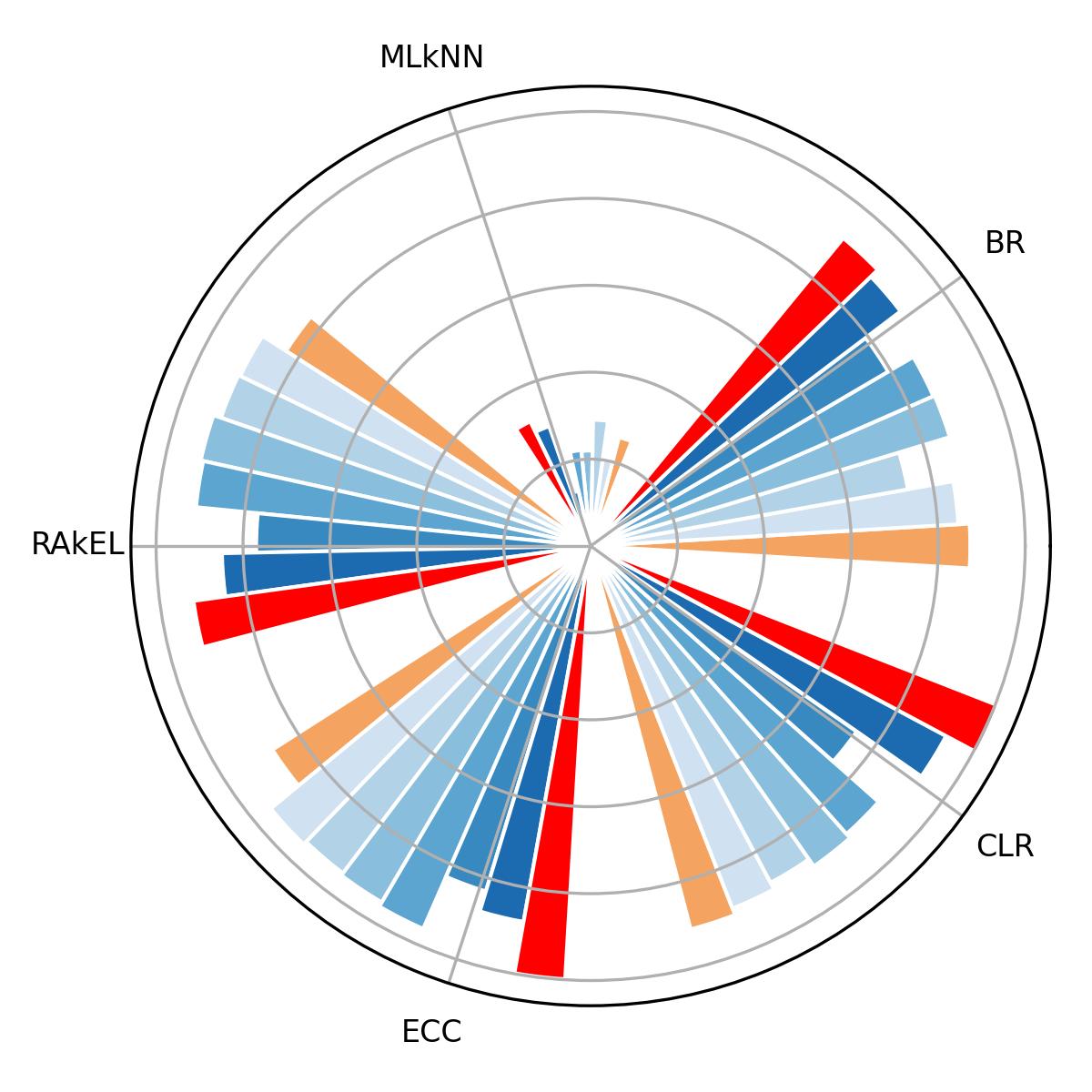}} 
\subfigure[rcv3]{\includegraphics[width=0.28\textwidth]{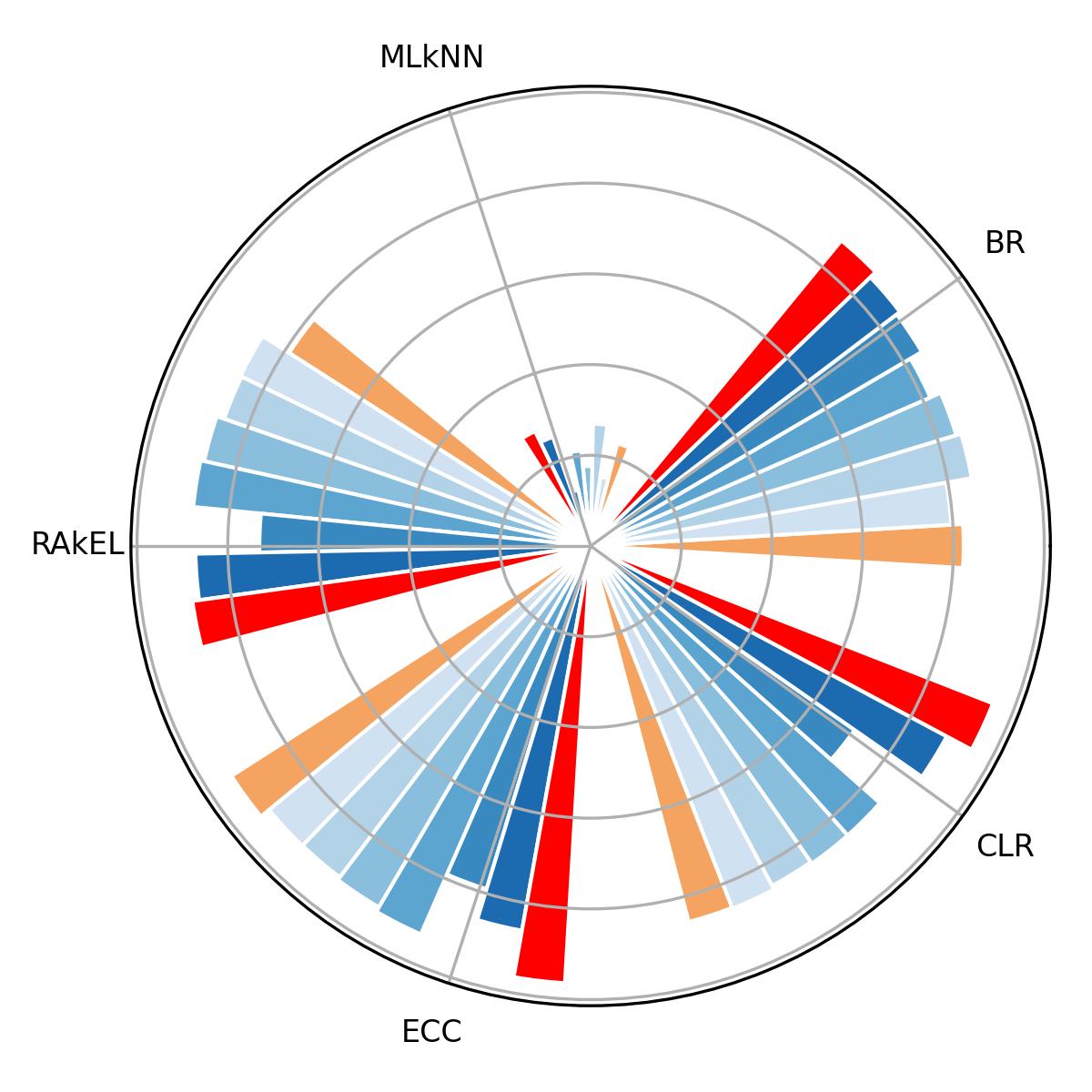}}
\subfigure[cal500]{\includegraphics[width=0.28\textwidth]{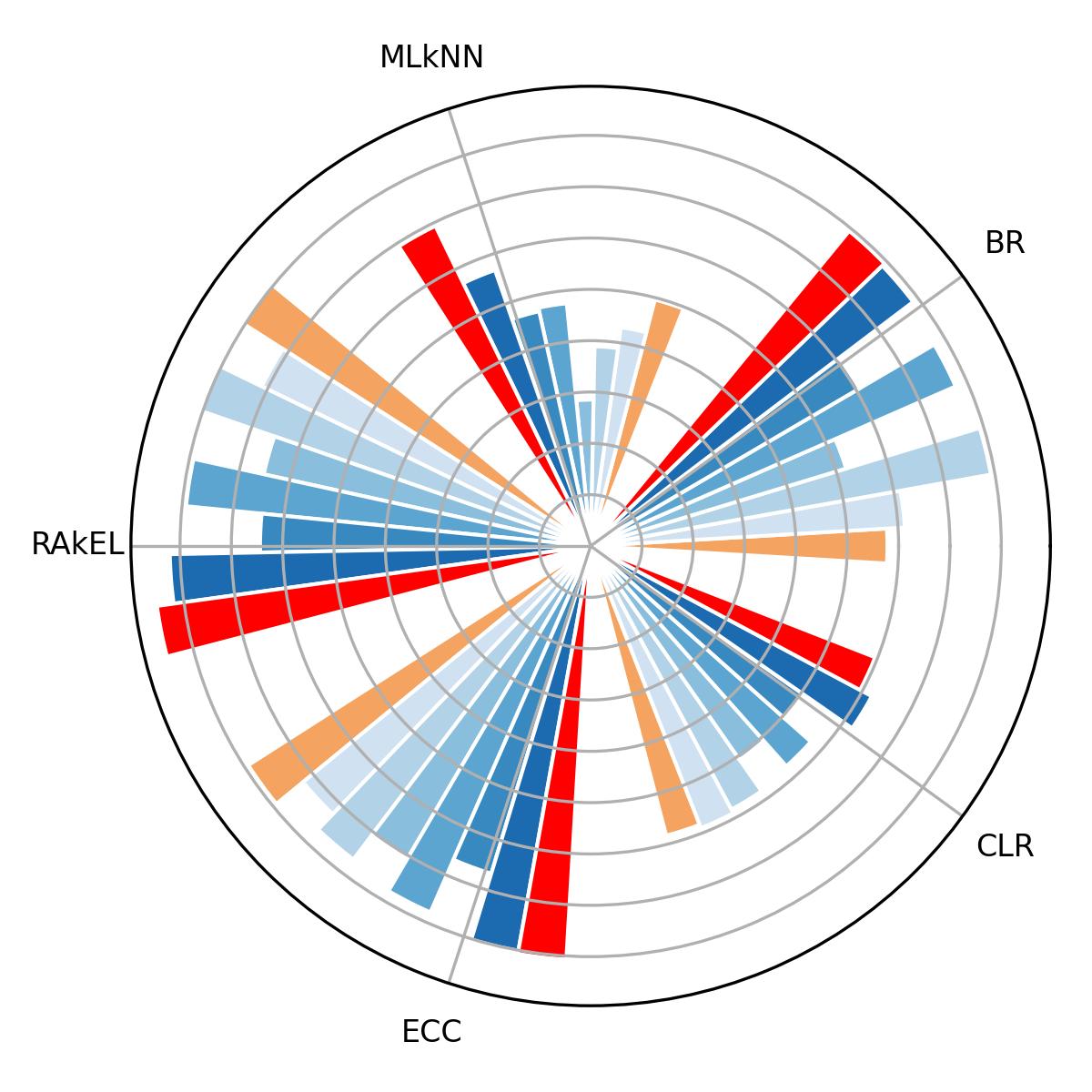}}
\subfigure[Corel5k]{\includegraphics[width=0.28\textwidth]{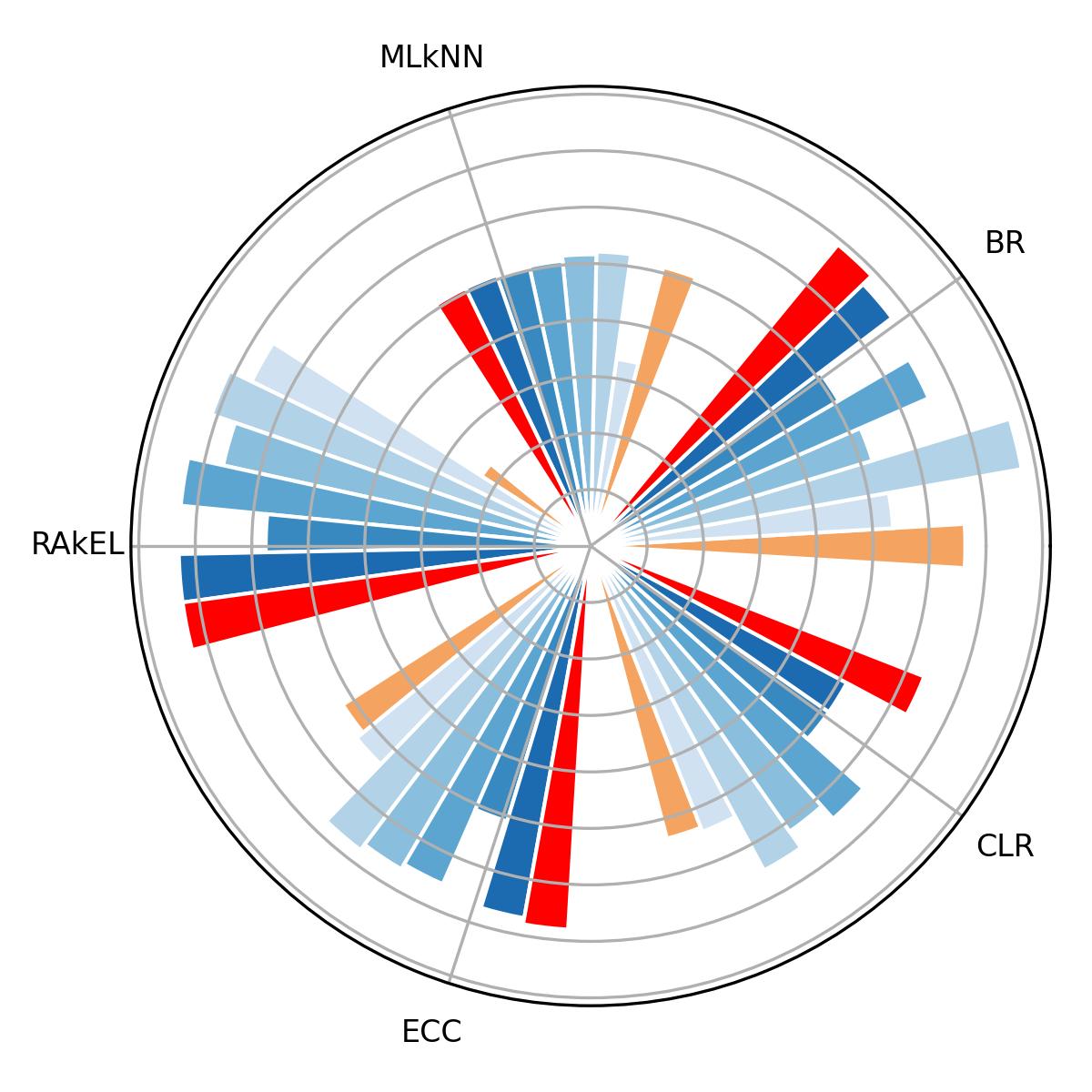}}
\subfigure{\includegraphics[width=0.9\textwidth]{legend.png}}
\caption{The performance of the multi-label sampling methods in terms of Ranking Loss across five different classification methods.}
\label{fig:Rankloss}
\end{figure}
Table~\ref{tab:RANK} presents the average rankings of each base classifier combined with sampling methods across all datasets. Additionally, the Friedman test was utilized to verify the significant superiority/inferiority of our method compared to other sampling approaches across three evaluation metrics in five basic multi-label classification methods. The detailed results of the comparative sampling methods using five fundamental learners on the Macro-F, Macro-AUC, and Ranking Loss are shown in Github. 
The Origin represents training directly using the training set without any sampling methods. 
The results indicate that the AEMLO method achieves the highest average ranking in almost all metrics, securing the most significant victories without any substantial losses. 
It is observed that MLBOTE and MLSOL outperform MLSMOTE, reflecting that refining rule selection for seed instances is more effective than oversampling with all minority seeds directly. An interesting observation is that the performance of MLTL and MLRUS is even worse than that of the original dataset. This is primarily attributed to the removal of critical instances, leading to the loss of important information. 

Autoencoders excel at generating new samples by learning compressed representations of input data. However, a subtle challenge arises when the feature space of these generated samples diverges from that of the origin samples. This divergence may pose difficulties for the MLkNN, which relies heavily on the distances between samples within the feature space to identify nearest neighbors. As such, any significant discrepancy in the feature distribution between generated and origin samples could potentially impact the MLkNN to accurately classify unseen samples. The enhanced performance of BR and CLR methods on augmented datasets can be attributed to the robustness of SVM and its adeptness at navigating complex decision boundaries. Specifically, SVM is particularly effective at managing the intricacies introduced into the feature space by data synthesized through Autoencoders.
\begin{table}[t]
\centering
\caption{The mean ranking of different sampling methods evaluated with five base learners across three metrics is presented. The notation (n1/n2) indicates adjustments from the Friedman test at a 5$\%$ significance level, signifying that the method in question significantly outperforms n1 methods and is outdone by n2 methods. The top-performing method is emphasized in bold, with lower rankings indicating superior performance.}
\label{tab:RANK}
\resizebox{\textwidth}{!}{
\begin{tabular}{cccccccccc}
\toprule
\multicolumn{1}{l}{} &  & Origin & MLSMOTE & MLSOL & MLROS & MLRUS & MLTL & MLBOTE & AEMLO \\ \midrule
\multirow{5}{*}{Macro-F} & BR & 6.33(0/3) & 5.11(0/3) & 2.89(4/1) & 4.22(1/0) & 6.00(0/3) & 7.33(0/5) & 2.78(4/0) & \textbf{1.33(6/0)} \\
& MLkNN & 5.33(0/4) & 3.11(3/0) & \textbf{1.56(4/0)} & 3.89(2/0) & 6.89(0/4) & 7.67(0/5) & 4.56(1/1) & 3.00(3/0) \\
& RAkEL & 5.67(0/3) & 4.11(2/2) & 2.67(2/0) & 4.89(2/2) & 7.22(0/5) & 7.00(0/5) & 2.44(3/0) & \textbf{2.00(4/0)} \\
& ECC & 5.11(0/2) & 4.22(0/0) & 3.56(2/0) & 5.67(0/2) & 5.44(0/2) & 6.67(0/3) & 2.78(3/0) & \textbf{2.56(4/0)} \\
& CLR & 6.33(0/3) & 3.89(2/0) & 2.78(3/0) & 3.00(2/0) & 6.89(0/4) & 7.11(0/5) & 3.89(2/0) & \textbf{2.11(3/0)} \\
& Avg(Total) & 5.75(0/15) & 4.09(7/5) & 2.69(15/1) & 4.33(7/4) & 6.49(0/18) & 7.16(0/23) & 3.29(13/1) & \textbf{2.20(20/0)} \\\midrule
\multirow{5}{*}{Macro-AUC} & BR & 5.00(0/2) & 4.78(0/1) & 3.00(3/0) & 4.11(1/0) & 5.78(0/3) & 7.56(0/5) & 4.33(1/1) & \textbf{1.44(7/0)} \\
& MLkNN & 5.11(0/3) & 4.22(2/0) & 3.44(3/0) & 3.78(2/0) & 6.11(0/4) & 6.78(0/4) & 4.44(1/0) & \textbf{2.11(3/0)} \\
& RAkEL & 5.33(0/2) & 4.33(1/0) & \textbf{2.78(2/0)}  & 3.56(1/0) & 4.33(1/0) & 7.67(0/6) & 4.89(0/0) & 3.11(2/0) \\
& ECC & 5.22(0/2) & 5.02(0/2) & 3.33(3/0) & 4.22(1/0) & 6.89(0/4) & 7.00(0/4) & 2.56(3/0) & \textbf{1.56(5/0)} \\
& CLR & 6.00(0/3) & 5.00(0/1) & 3.16(2/0) & 3.67(2/1) & 5.56(0/2) & 7.44(0/4) & 3.33(2/0) & \textbf{1.67(5/0)} \\
& Avg(Total) & 5.33(0/12) & 4.71(3/4) & 2.98(13/0) & 3.87(7/0) & 5.73(1/13) & 7.29(0/23) & 3.91(7/1) & \textbf{2.18(22/0)} \\\midrule
\multirow{5}{*}{Ranking Loss} & BR & 4.89(0/1) & 6.22(0/2) & 4.44(0/0) & 4.00(1/0) & 5.33(0/2) & 6.11(0/3) & 3.00(3/0) & \textbf{2.00(4/0)} \\
& MLkNN & 4.56(0/2) & 5.44(0/3) & 3.56(3/0) & 5.33(0/2) & 4.00(0/2) & 6.89(0/5) & \textbf{2.89(3/0)} & 3.33(2/0) \\
& RAkEL & 6.44(0/3) & 5.22(0/1) & 4.33(1/0) & 4.44(0/0) & 2.78(2/0) & 7.56(0/4) & 3.11(2/0) & \textbf{2.11(3/0)} \\
& ECC & 5.00(0/1) & 4.56(0/0) & 4.78(0/0) & 3.67(1/0) & 4.67(0/0) & 6.33(0/2) & 4.11(0/0) & \textbf{2.89(2/0)} \\
& CLR & 5.22(0/2) & 4.00(1/1) & 4.22(1/1) & 4.11(1/0) & 5.00(0/2) & 7.67(0/5) & 3.89(3/0) & \textbf{1.89(5/0)} \\ 
& Avg(Total)&5.22(0/9) & 5.09(1/7) & 4.27(5/1) & 4.31(3/2) & 4.36(2/6) & 6.91(0/19) & 3.40(11/0) & \textbf{2.44(16/0)} \\\bottomrule
\end{tabular}
}

\end{table}

\subsection{Parameter Analysis}
We investigate the influence of various parameter settings on the performance of ALMLO. We select smaller enron and larger Corel5k as two representative datasets in the parameter analysis.

\begin{figure}[h]
\centering
\subfigure[enron]{\includegraphics[width=0.45\textwidth]{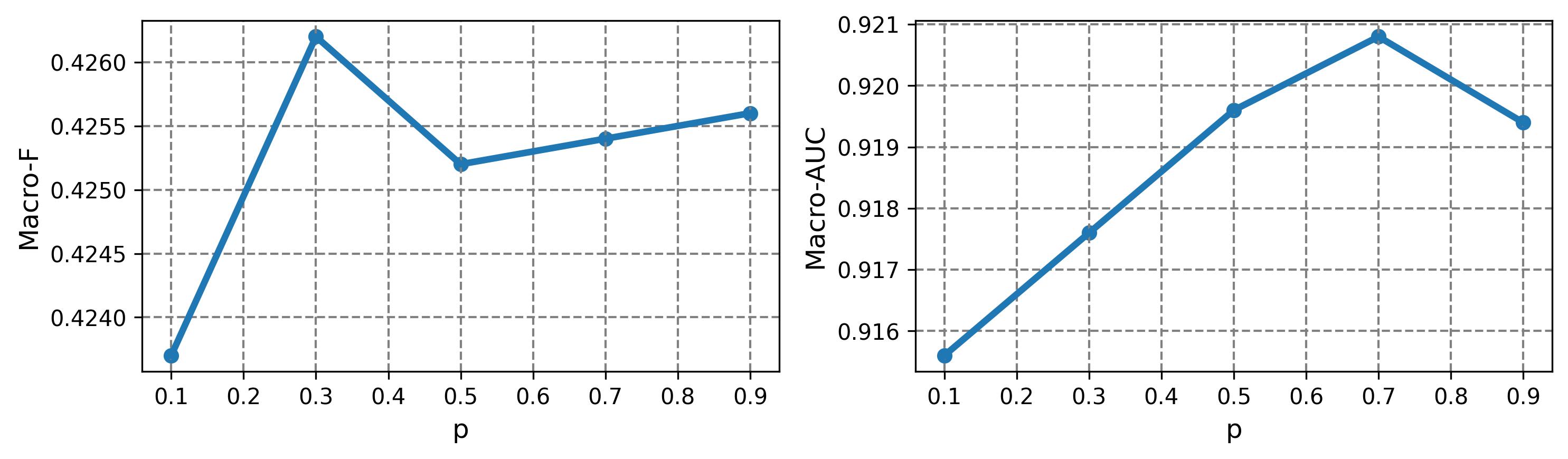}
\label{fig:penron}}
\subfigure[Corel5k]{\includegraphics[width=0.45\textwidth]{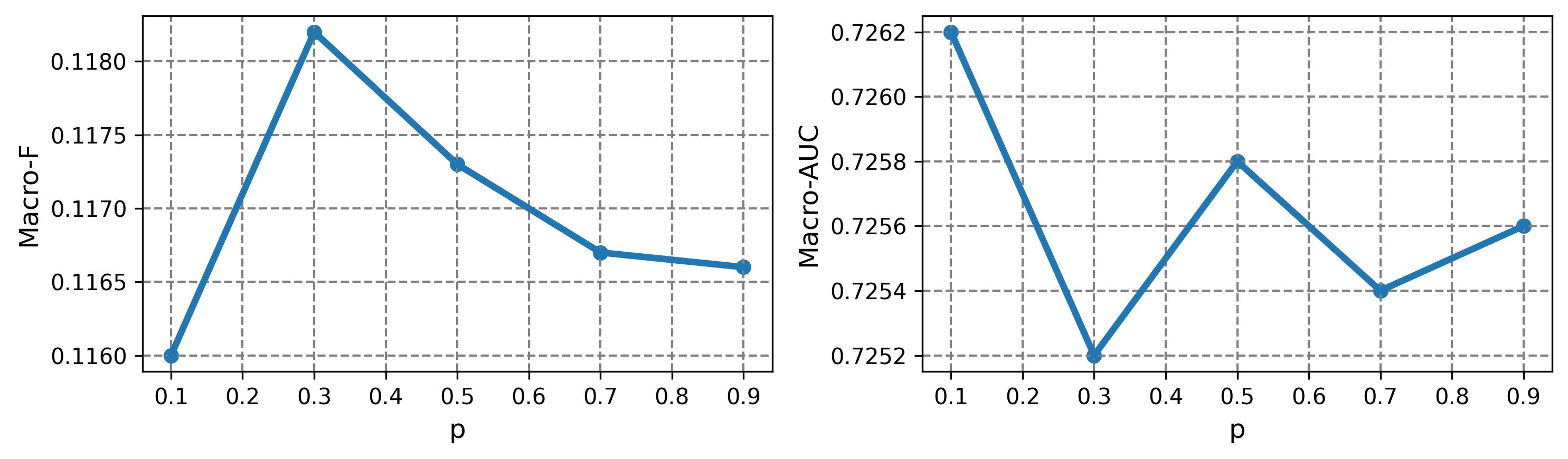}
\label{fig:pCorel5k}}
\caption{Performance of sampling rate in terms of Macro-F and Macro-AUC.}
\label{fig:p}
\end{figure}

As shown in Figure \ref{fig:penron}, the impact of varying sampling rate \( p \) on Macro-F and Macro-AUC scores (based on MLkNN) shows a trend of initial fluctuation, followed by stabilization. 
In contrast, in Figure \ref{fig:pCorel5k} exhibits a higher sensitivity to \( p \), with significant volatility in Macro-F and inconsistent variations in Macro-AUC. These observations suggest that the optimal selection of \( p \) may be highly dependent on dataset characteristics. 

\begin{figure}[!h]
\centering
\subfigure[enron]{\includegraphics[width=0.45\textwidth]{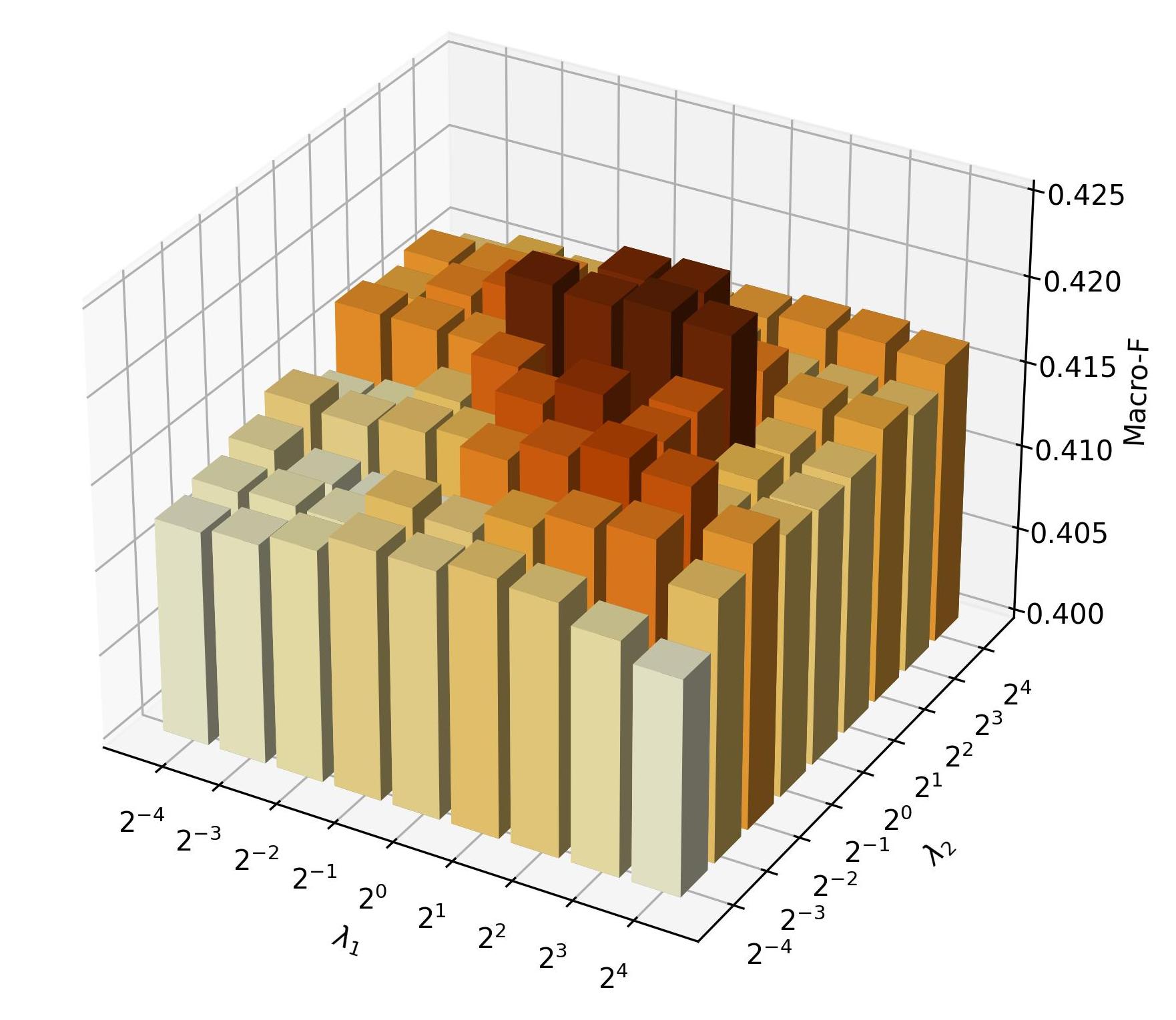}}
\subfigure[Corel5k]{\includegraphics[width=0.45\textwidth]{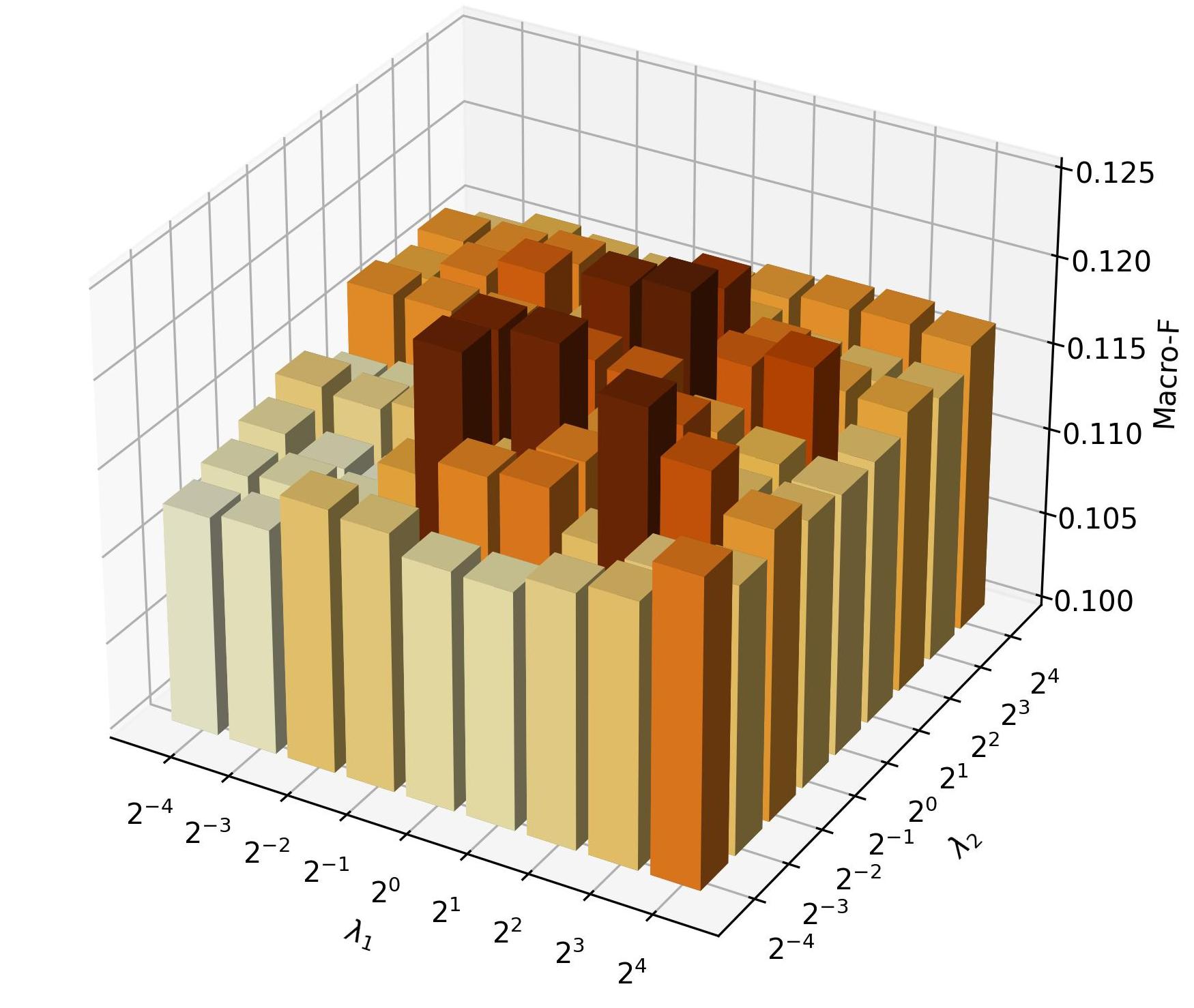}}
\caption{Performance of AEMLO with varying parameter configurations in terms of Macro-F.}
\label{fig:alphabeta}
\end{figure}

Figure \ref{fig:alphabeta} illustrates ALMLO's performance sensitivity to variations in \(\alpha\) and \(\beta\), highlighting the importance of balancing feature reconstruction loss with label relevance loss during optimization. 

\subsection{Sampling Time}
\begin{figure}[!h]
\centering
\subfigure[enron]{\includegraphics[width=0.45\textwidth]{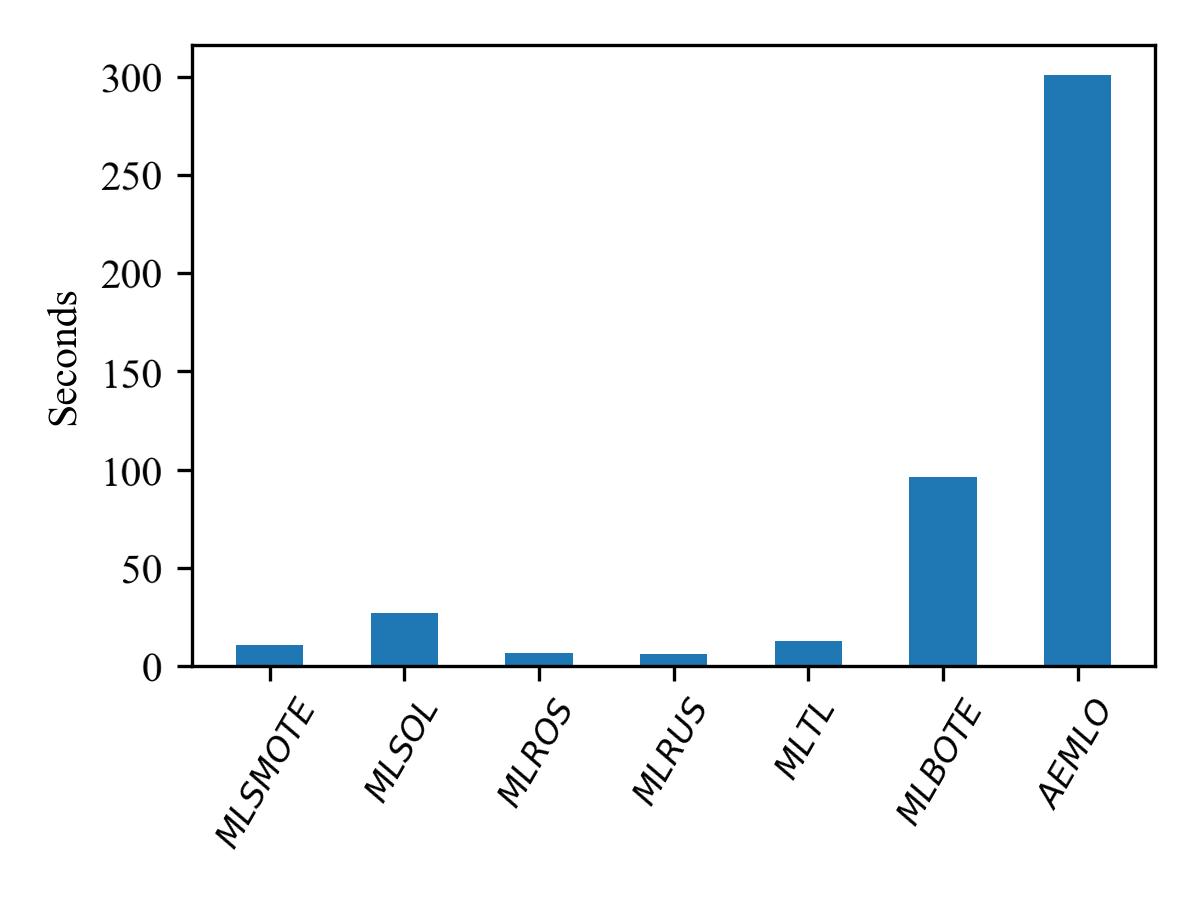}}
\subfigure[Corel5k]{\includegraphics[width=0.45\textwidth]{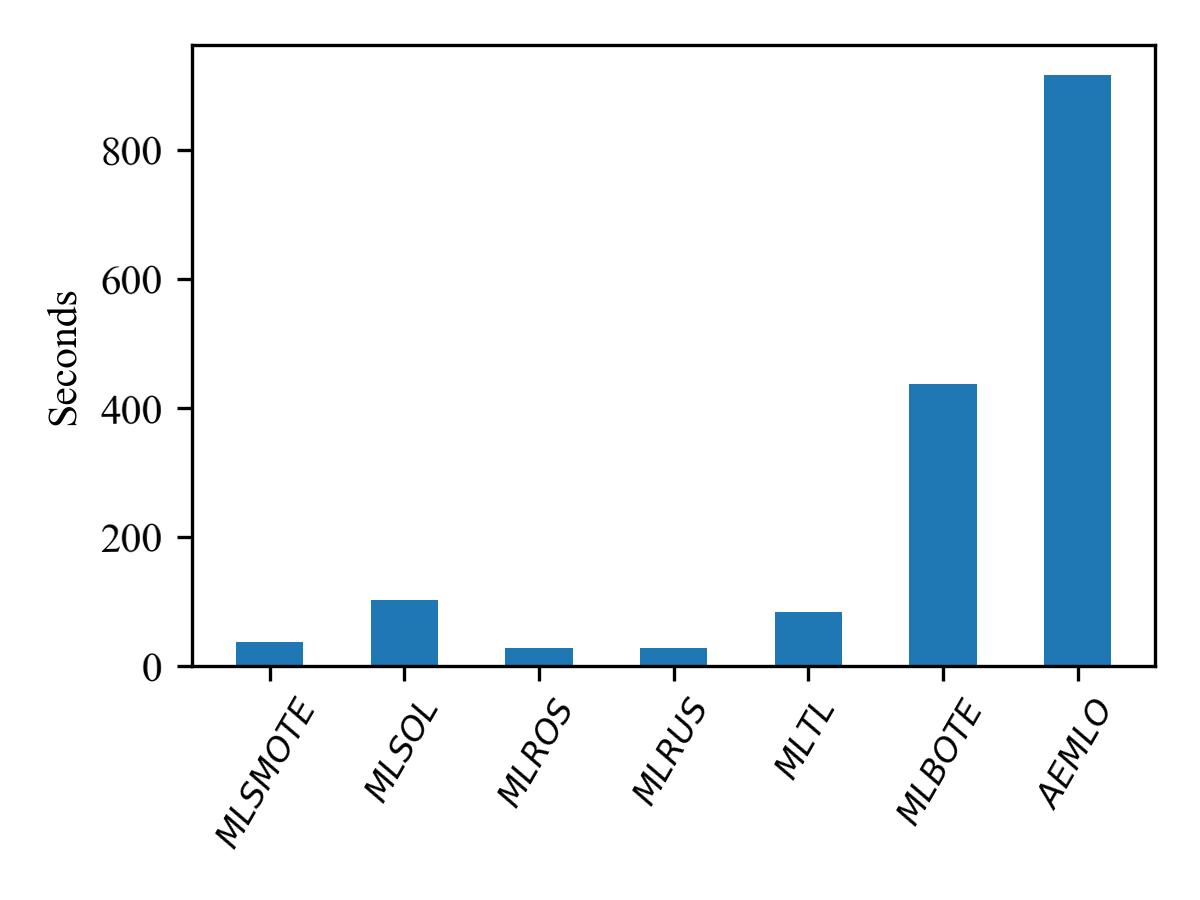}}
\caption{Sampling time of different sampling method.}
\label{fig:time}
\end{figure}

Figure \ref{fig:time} shows the time efficiency of different sampling methods, with the epoch set as 100 for AEMLO. It is evident that AEMLO, as a deep learning approach, requires training before sampling, resulting in a higher time expenditure. 

\section{Conclusion}
In this paper, we introduce AEMLO, an innovative oversampling model devised for addressing data imbalance in multi-label learning by integrating canonical correlation analysis with the encoder-decoder paradigm. AEMLO emerges as an effective oversampling solution for training deep architectures on imbalanced data distributions. It acts as a data-level solution for class imbalance, synthesizing instances to balance the training set and thus enabling the training of any classifiers without bias. AEMLO exhibits the pivotal characteristics crucial for a successful sampling algorithm in the multi-label learning domain: the ability to manipulate features and labels, i.e., to learn low-dimensional joint embeddings from feature and label representations and transform them into an original-dimensional space, along with generating new feature representations and their corresponding label subsets. This is facilitated through the utilization of an encoder/decoder framework. Extensive experimental studies demonstrate the capability of AEMLO to handle imbalanced multi-label datasets in various domains and collaborate with diverse multi-label classifiers. 



\subsubsection{Acknowledgements.} This work was supported by the National Natural Science Foundation of China (62302074) and the Science and Technology Research Program of Chongqing Municipal Education Commission (KJQN202300631).
%
%
%
%

\end{document}